\pdfoutput=1
\documentclass[11pt]{article}

\usepackage[]{ACL2023}

\definecolor{dkgreen}{RGB}{0,179,36}

\usepackage{lipsum}

\usepackage{times}
\usepackage{latexsym}
\usepackage{subcaption}
\usepackage[T1]{fontenc}
\usepackage{amsfonts, amsmath}
\usepackage{enumitem}
\usepackage[utf8]{inputenc}

\usepackage{microtype}

\usepackage{inconsolata}

\usepackage{xcolor}
\usepackage{booktabs, multirow} % for borders and merged ranges
\usepackage{makecell}
\usepackage{soul}% for underlines
\usepackage{changepage,threeparttable} % for wide tables
\usepackage{graphicx} % for figure
\usepackage{xurl}
\title{SLUE Phase-2:  A Benchmark Suite of Diverse \\ Spoken Language Understanding Tasks}

\author{Suwon Shon$^{1}$, Siddhant Arora$^{2*}$, Chyi-Jiunn Lin$^{3*}$, Ankita Pasad$^{4}$\thanks{\;\;Core contributors in alphabetical order}\;\;,\\ {\bf Felix Wu}$^{1}$, {\bf Roshan Sharma}$^{2}$,{\bf Wei-Lun Wu}$^{3}$,\\ {\bf Hung-Yi Lee}$^{3}$, {\bf Karen Livescu}$^{4}$, {\bf Shinji Watanabe}$^{2}$ \vspace{4px} \\ 
 $^1$ASAPP \ \ \ $^2$Carnegie Mellon University \ \ $^3$National Taiwan University \\ $^4$Toyota Technological Institute at Chicago \\}

\begin{document}
\maketitle
\begin{abstract}
Spoken language understanding (SLU) tasks have been studied for many decades in the speech research community, but have not received as much attention as lower-level tasks like speech and speaker recognition. 
In this work, we introduce several new annotated SLU benchmark tasks based on freely available speech data, which complement existing benchmarks and address gaps in the SLU evaluation landscape.  We contribute four tasks:  {\it question answering} and {\it summarization} involve inference over longer speech sequences; {\it named entity localization} addresses the speech-specific task of locating the targeted content in the signal; {\it dialog act classification} identifies the function of a given speech utterance.  
In order to facilitate the development of SLU models that leverage the success of pre-trained speech representations, we will 
release a new benchmark suite, including for each task (i) curated annotations
for a relatively small fine-tuning set, 
(ii) 
reproducible pipeline (speech recognizer + text model) and end-to-end baseline models and evaluation metrics, (iii) baseline model performance in various types of systems for easy  comparisons. 
We present the details of data collection and annotation and the performance of the baseline models. 
We also analyze the sensitivity of pipeline models' performance 
to the speech recognition accuracy, using more than 20 publicly available
speech recognition models.

\end{abstract}

\section{Introduction} \label{sec:introduction}
Spoken language understanding (SLU) tasks involve inferring the linguistic structure or semantic meaning of a speech signal beyond its text transcript.
We use this term broadly to include any natural language processing (NLP) task applied to speech, and tasks that involve  linguistic understanding but also localization in the signal of relevant segments or producing speech as output. SLU has been an active area throughout the history of speech research~\cite{hemphill1990atis,calhoun2010nxt,busso2008iemocap,zadeh2018multimodal,chen2020large,cohn2019audio,yadav2020end,martinez2020msp}.  However, compared to "lower-level" tasks like automatic speech recognition (ASR) and speaker identification, SLU has received much less attention and resources, and specifically there are much fewer benchmarks with freely available data.

SLU tasks can in principle be addressed via a pipeline approach --- using ASR to map speech to text and an NLP (text) model to map text to the desired output.  The alternative is an end-to-end (E2E) model, which maps directly from the input speech to the target output. While pipeline approaches can take advantage of existing strong ASR and NLP models, E2E models can be more efficient at inference time, can avoid ASR error propagation, and can directly use aspects of the speech signal beyond the text that are useful for the task (e.g., prosody)~\citep{prosody-useful,chen2020acoustic,jurafsky1998lexical,tran2018parsing}.  In addition, for tasks whose output includes speech segments or time spans, there is no direct combination of an ASR model and an NLP model that produces precisely the desired type of output.
For some SLU tasks, the current state of the art is a pipeline model~\cite{shon2022slue,Lai2020SemiSupervisedSL}, whereas for others E2E models are better~\cite{Pasad2021OnTU,sharma2022end,wu2022wav2seq,pretrain-E2E-SLU,arora2022espnet,Shon2022ContextawareFO}.  In order to better understand the pros and cons of pipeline and E2E approaches, more public benchmarks are sorely needed.

While collecting large amounts of labeled speech data for many SLU tasks may be prohibitively costly, recent advances in pre-trained models~\citep{baevski2020wav2vec,hsu2021hubert,Chen2021WavLMLS,wu2022performance,Baevski2022data2vecAG,lin2022melhubert,mohamed2022self} make it feasible to use relatively small fine-tuning sets for each task.  There have been several recent efforts to introduce new benchmark SLU tasks~\cite{yang2021superb,bastianelli2020slurp,feng2021asrglue,evain2021lebenchmark,arora2022espnet,lugosch2021timers,shon2022slue,tomasello2022stop}, most (but not all) using fairly small training sets of several hours to several dozens of hours of speech. 
Among them, the Spoken Language Understanding Evaluation (SLUE)\footnote{We refer to the original SLUE as "SLUE Phase-1."}~\cite{shon2022slue} motivated us since it pursues a natural speech, rather than a short command type of speech that is populated in other benchmarks. However, there are only two SLUE tasks (sentiment analysis and named entity recognition), thus more tasks with different complexities are needed to cover the diverse application of SLU.

% Among them, we motivated from the 
% % In this work, we motivated from 
% % In this work, we take as a starting point 
% the Spoken Language Understanding Evaluation (SLUE)\footnote{We refer to the original SLUE as "SLUE Phase-1."} benchmark suite~\cite{shon2022slue}. The benchmark includes two SLU tasks (sentiment analysis and named entity recognition) in addition to ASR; however, more tasks with different complexities are needed to cover the diverse applications of SLU.

% which has several advantages compared to other recent work in section~\ref{sec:related_work}:

% 1) high-quality human annotations, 2) Creative Commons (CC) licensed data and annotation, and 3) natural speech data as opposed to synthesized speech.  
% (TODO: do we want to mention natural vs. {\bf read} speech also?  read speech may be less likely to include features that E2E models can take advantage of; but our new data is a mix of read and not-read speech) 
% The existing SLUE benchmark includes two SLU tasks (sentiment analysis and named entity recognition) in addition to ASR; however, more tasks with different complexities are needed to cover the diverse applications of SLU.
%and applications are needed. 

We introduce SLUE Phase-2, a set of SLU tasks that complement the existing SLU datasets or benchmarks.  The new tasks include dialog act classification (DAC), question answering (QA), summarization (SUMM), and named entity localization (NEL)
, applied to English speech data. 
% Rather than expanding language, we keep the single language setup while focusing on diverse tasks with various domains. 
SLUE Phase-2 has several advantages compared to other recent work introduced in section~\ref{sec:related_work}:

\noindent
\textbf{More diverse tasks}: SLUE phase-2 not only include utterance or word-level classification task but also includes QA and SUMM task. 

\noindent
\textbf{More challenging tasks}: The complexity of the task is influenced by the type of input and the type of output. SLUE phase-2 uses conversational or longer discourse speech as input. The type of output is not limited to labels or text, but also includes the speech span time stamp.

\noindent
\textbf{New human annotation}: A new annotation was collected by a human annotator. Human annotator validated an automatically-collected data if needed.

\noindent
\textbf{Natural speech}: We do not use synthesized speech. We only include conversational or considerably long discourse speech rather than short speech commands.

\noindent
\textbf{CC license}: Creative Common licensed dataset to give the best freedom of use.

% These tasks have varying levels of complexity. Task complexity is influenced by the type of input (single utterance, longer discourse), the type of output (single discrete label, sequence of discrete labels, entity level time-stamps), and the nature of the mapping between input and output.
% These tasks have important applications and are also diverse in terms of their complexity, type of input (single utterances, longer discourse), and type of output (discrete labels, time stamps).  

For each task, we provide publicly available\footnote{To be released.} datasets, annotations, models, and code. We provide both pipeline and E2E baseline models and, for pipeline models, we use multiple ASR systems to analyze the effect of the ASR error rate on the final task performance.

\section{Related work} \label{sec:related_work}

% \subsection{SLU datasets and tasks}
\textbf{SUPERB}~\citep{yang2021superb} aggregates several existing speech tasks mainly to evaluate frozen pre-trained speech models. 
It focuses on low-level tasks but also contains two SLU tasks --- intent classification (from Fluent Speech Commands~\citep{lugosch2019speech}) and slot filling (from SNIPS~\citep{coucke2018snips}).
However, the former is an easy task where many models have close to 100\% accuracy, and the latter uses synthesized rather than natural speech.
\textbf{SLURP}~\citep{bastianelli2020slurp} is a spoken version of a text dataset~\citep{liu2019benchmarking} where the authors hired workers to dictate the written conversations between humans and personal robot assistants. It includes three SLU tasks --- scenario prediction, action prediction, and entity prediction. These tasks cannot be generalized as the nature of the short speech command.
\textbf{ASR-GLUE}~\citep{feng2021asrglue} is based on the well-known GLUE benchmark~\citep{wang2018glue} where the authors hired people to speak the GLUE text 
% with three levels of background noise
. It includes five GLUE tasks and one additional task.
%It includes 6 tasks: SST-2~\citep{Lee2015SentimentAS}, STS-B~\citep{Cer2017SemEval2017T1}, QQP\footnote{\url{https://quoradata.quora.com/First-Quora-Dataset-Release-Question-Pairs}}, QNLI~\citep{Rajpurkar2016SQuAD1Q}, RTE~\citep{BarHaim2006TheSP,Dagan2007ThePR}, SciTail~(\citet{Khot2018SciTaiLAT}; the last one is not part of GLUE). Notab
However, ASR-GLUE contains only a test set; researchers must rely on other datasets for training.
\textbf{Timers and Such}~\citep{Lugosch2021TimersAS} is a dataset of speech commands that involve numbers, designed for intent classification and slot filling that has limited use case.
\textbf{Spoken SQuAD}~\citep{lee2018spoken} and \textbf{Spoken CoQA}~\citep{you2022end} are synthesized speech versions of the text SQuAD~\citep{Rajpurkar2016SQuAD1Q} and CoQA~\citep{reddy2019coqa} datasets.
\textbf{NMSQA}~\citep{lin2022dual} is a multi-speaker spoken QA dataset whose test set contains natural speech but the train and validation sets are synthesized. 
Other well-known SLU datasets include \textbf{ATIS}~\citep{hemphill1990atis} and \textbf{Switchboard NXT}~\citep{calhoun2010nxt}, which have been used for tasks like intent and DAC, but the data is available under license constraints. 
~\citet{wu2020harpervalleybank} published an open-sourced speech dataset; however, its dialog act annotation is not manually annotated but predicted using commercial API. 

%% Added by Roshan- previous data for summarization

Speech summarization has gained interest over the past few years with tasks such as abstractive summarization of instructional \textbf{How-2} videos ~\citep{sanabria2018how2} and \textbf{TED Talks} ~\citep{attention-fusion}, but the raw audio for these tasks is not publicly available. Other corpora, such as the \textbf{ICSI} ~\citep{janin2003icsi} and \textbf{AMI} ~\citep{mccowan2005ami} meeting summarization corpora, contain relatively less annotated data. 
Named entity localization (NEL) is a fairly new task. A similar task, audio de-identification (audio de-ID), has been introduced with annotations for conversational data from Switchboard and Fisher~\citep{cohn2019audio, baril2022named},  but these datasets are not free.
%but has not been explored thoroughly. \citet{cohn2019audio} propose a pipeline approach -- speech-to-text followed by text NER and speech-text forced alignment module -- for audio de-ID. \citet{baril2022named} simplify the task further by using the ground-truth transcripts for a pipeline approach. 
Audio de-ID is a special case of NEL where the entities of interest are related to personal identifiers. 
%To the best of our knowledge, ours is the first work to introduce an end-to-end (E2E) baseline for the task.

We focus on English speech-related work (most comparable with our work), but there are also ongoing efforts for other languages~\citep{tomashenko2019recent,evain2021lebenchmark}.

\section{SLUE Phase-2:  Tasks and data} \label{sec:slue-2}
This section introduces the tasks and metrics in SLUE Phase-2. The SLUE phase-1 introduced the "SLUE score", a numerical summary of model performance across tasks. However, as we consider a more diverse set of tasks, using the same pre-trained model for all tasks is difficult, and evaluation via a single SLUE score may discourage building systems for individual tasks.  In SLUE Phase-2, therefore, we do not adopt the single SLUE score, and evaluate each task individually.
%could make researchers hesitant to submit their results since they have to develop models for all the SLUE tasks. Thus we would no longer use the SLUE score and researchers could treat each SLUE task as an individual task.

\subsection{Tasks}
We explore more diverse and complex tasks compared to SLUE phase-1. As an extension of NER task in SLUE, we describe the NEL task to predict the audio time-stamps of named entities. DAC is an utterance classification task within conversation interactions to predict dialog acts given input speech. We address two longer-range context tasks: QA and SUMM where the model takes a long sequence and utilizes context across the entire scope to answer questions or summarize speech respectively. 

% SLUE Phase-1 explored tasks that predict named entities or sentiment from speech, and Phase-2 expands this by considering more complex and diverse tasks. As an extension of Named Entity Recognition, we consider the perhaps more challenging task of predicting the audio time-stamps of named entities using Named Entity Localization (NEL). Dialog Act Classification (DAC) is an utterance-level classification task like sentiment, but is perhaps more challenging because of conversational interactions inherent in dialog, and the lack of punctuation and casing in typical speech recognizers. We address two longer range context tasks- Spoken Question Answering (QA) and Speech Summarization(SUMM), where models take as input longer speech sequences and utilize context across the entire scope to answer questions or summarize speech respectively. 
\subsubsection{Dialog Act Classification (DAC)} 
DAC is the task of identifying the function of a given speech utterance in a dialog, such as question, statement or backchannel.
It is an utterance-level multi-label multi-class classification task; that is, an utterance can have more than one class (function).
We evaluate DAC using macro-averaged (unweighted) F1 score.
\subsubsection{Question Answering (QA)} 
The goal of QA is to find the answer span in a spoken document given a spoken question. The answer span is denoted by the start and end frames of a short phrase in the document. We use the frame-level F1 (frame-F1) score~\citep{chuang2020speechbert} to evaluate the overlap between the predicted and the ground-truth answer spans.
\subsubsection{Speech summarization (SUMM)} 
SUMM refers to the task of generating a text summary from a given speech input. SUMM is challenging as it requires a model to assimilate information across very long input contexts in order to identify essential information and paraphrase to obtain the abstractive summary of speech. 
% processing of long speech inputs consisting of a sequence of spoken utterances to generate a concise representation of the audio. In this work, we focus on abstractive summarization, i.e., the summary is generated by paraphrasing the important information from the speech document. 
We evaluate SUMM using ROUGE~\cite{lin-2004-rouge}, METEOR~\cite{denkowski-lavie-2014-meteor} and BERTScore~\cite{bert-score}.
\subsubsection{Named Entity Localization (NEL)}
The goal of NEL is to predict the start and end times of any named entities in a spoken utterance.
% Named entities are phrases, often consisting of proper nouns, that refer to distinct entities, e.g.~a person, location, organization, numerical value.
NEL is related to named entity recognition (NER), but NER involves identifying entity phrases while NEL involves locating them in the audio.
%traditionally, NER predicts (text) entity phrases and corresponding entity labels. NEL locates the entity phrases in the input audio. 
We evaluate performance via two F1 scores based on the overlap between the predicted and ground-truth time ranges: {\it frame-F1}, defined similarly to the QA {\it frame-F1} measure; and {\it word-F1}, defined similarly to the de-identification metric of \citet{cohn2019audio}.  The {\it word-F1} score has a hyperparameter $\rho \in [0,1]$, which is the fraction of overlap between a ground-truth word segment and the predicted region needed to count the word as detected; $\rho=1$ means a perfect match is required.
%for a prediction to count as a hit.
%The overlap is measured either as (i) the number of frames, measured similarly to QA's {\it frame-F1} measure, or (ii) the number of words, {\it word-F1},  defined similarly to \citet{cohn2019audio} and has a hyperparameter $\rho$. $\rho$ ($\leq1$) is the fraction of overlap between a ground-truth word segment and the predicted region to count as one word, so, $\rho=1$ means that a perfect match is required for the prediction to count as a hit or a miss.

% NEL is evaluated via overlap between the predicted and the ground-truth entity regions, quantified either as the number of frames or the number of words, reported as {\it frame-F1} or {\it word-F1} respectively. The {\it word-F1} score is defined similarly to \citet{cohn2019audio} with a hyperparameter $\rho$ to allow tolerance for the overlap. $\rho$ ($\leq1$) is the fraction of overlap with a ground-truth word segment for the predicted region to count as one word, so, $\rho=1$ means a perfect match is required for the prediction to count as a hit or a miss. The {\it frame-F1} score does not have any hyperparameters.

\subsection{Datasets and annotation}
\subsubsection{SLUE-HVB for DAC}
For the DAC task we adapt the Harper Valley Bank (HVB) spoken dialog corpus\footnote{\scriptsize{CC-BY-4.0 license}}~\cite{wu2020harpervalleybank}
of scripted consumer banking dialogs, simulated by 59 
speakers.
The data contains about 23 hours of audio from 1,446 conversations with transcriptions and metadata, as well as dialog act annotation. 
However, the original DAC annotation is automatic, without manual validation, and the set of dialog acts is simple and tailored to this corpus.
We define a new set of acts and collect human annotations by professional annotators listening to the audio.  Our new set of dialog acts (See Table~\ref{tab:da_detail} in Appendix for detail) is based on the well-known Switchboard NXT~\cite{calhoun2010nxt} dialog act set.  Based on a pilot annotation, we remove several unneeded labels and merge others unnecessarily granular.
%To define a new set of acts, we first use the Switchboard NXT~\cite{calhoun2010nxt} dialog act set to annotate a few conversations from the HVB dataset. From this pilot annotation, we find that many acts are not applicable or unnecessarily granular. Thus we modify the dialog acts set by removing and merging, and finalize a set of 18 acts. The detail act list is 
%The final label set is shown in Table~\ref{tab:da_detail} in Appendix. 
%Then, HVB was annotated by human annotators listening to audio. 
%For benchmark evaluation, w
Finally, we split the HVB data into fine-tune, dev, and test sets (Table~\ref{tab:hvb_stats}). The intent of conversation is balanced along the splits. We exclude short audio clips (<210ms) and audio that contains no speech.
%doesn't have speech.
% since this benchmark data will be used to evaluate the E2E system and pipeline system as well.

% \begin{table}[!htp]\centering
% \caption{Dialog acts list (see appendix for detail).}\label{tab:acts_set}
% \resizebox{5cm}{!}{%
% \begin{tabular}{c|c}\toprule
% {actions} &{sub-actions} \\\midrule
% \multirow{3}{*}{question} &{question\_check} \\
% &{question\_repeat} \\
% &{question\_general} \\ \midrule
% \multirow{3}{*}{answer} &answer\_agree \\
% &{answer\_dis} \\
% &{answer\_general} \\ \midrule
% \multirow{8}{*}{statement} & apology \\
% &{thanks} \\ 
% &{acknowledge} \\
% &{statement\_open} \\
% &{statement\_close} \\
% &{statement\_problem} \\
% &{statement\_instruct} \\
% &{statement\_general} \\ \midrule
% \multirow{3}{*}{natural speech} &{backchannel} \\
% &{disfluency} \\
% &{self} \\ \midrule
% other &{other} \\
% \bottomrule
% \end{tabular}
% }
% \end{table}

\begin{table}[!htp]\centering
\caption{SLUE-HVB data statistics}\label{tab:hvb_stats}
\resizebox{5cm}{!}{%
\begin{tabular}{l|r|r}\toprule
&utterances &duration (h) \\\midrule
fine-tune &11,344 &6.8 \\
dev &1,690 &1.0 \\
test &6,121 &3.6 \\
\bottomrule
\end{tabular}
}
\end{table}
\subsubsection{SLUE-SQA-5 for QA}
\label{sssection:qa-dataset}

%Most of the past open-source English spoken QA datasets, including Spoken SQuAD~\citep{lee2018spoken}, NMSQA~\citep{lin2022dual}, Spoken-CoQA~\citep{you2022end}, do not have a large training set consisting of realistic human speech.
%Instead, these three datasets use text-to-speech systems to generate synthesized speech for their training sets.
%However, experiments in \citet{lin2022dual} and \citet{you2022end} showed that QA models trained with synthesized speech might have bad performance on real speech test set despite good performance on the synthesized speech test set, and it remains unclear whether these QA models can have higher accuracy when trained or fine-tuned with real speech.
%As a result, in this paper, we propose a new spoken QA dataset, SLUE-SQA-5, whose fine-tune, dev, and test set all consists of real speech QA data.

Previous open-source English spoken QA datasets, including Spoken SQuAD~\citep{lee2018spoken}, NMSQA~\citep{lin2022dual}, Spoken-CoQA~\citep{you2022end}, do not have a large training set consisting of realistic human speech, so we propose a new spoken QA dataset, SLUE-SQA-5, whose fine-tune, dev, and test sets all consist of real speech data. %for fine-tuning and evaluating spoken QA models.

The text transcriptions of question-answer pairs in SLUE-SQA-5 are collected from five different text QA datasets: SQuAD\footnote{\scriptsize{CC BY-SA 4.0 license}}~\citep{Rajpurkar2016SQuAD1Q}, Natural Questions\footnote{\scriptsize{CC BY-SA 3.0 license}} (NQ)~\citep{kwiatkowski2019natural}, TriviaQA\footnote{\scriptsize{Apache License 2.0}}~\citep{joshi2017triviaqa}, WebQuestions\footnote{\scriptsize{CC-BY 4.0 license}} (WQ)~\citep{berant2013semantic}, 
 and CuratedTREC
%$^6$
  \footnote{\scriptsize{Public Domain}}
(TREC)~\citep{baudivs2015modeling}. 
 %These five text QA datasets have mined their question-answer pairs in different ways and thus their questions have different distributions of domains, question types, and sentence lengths. 
We gather the text questions from the training set of the five text QA datasets as our fine-tune set.
For our dev and test sets, we first collect the questions from the dev set of SQuAD, NQ, TriviaQA, WQ and the test set of TREC, and then randomly split these questions into two subsets as our dev and test sets. 
 To get the spoken version of the collected questions, we used Amazon Mechanical Turk (MTurk), a crowdsourcing platform with anonymous, non-expert workers, to collect spoken questions read by human speakers.
The collection details are shown in Section~\ref{ssec:question-collection} in the Appendix.

% , since the total number of words of the documents is much higher than that of the questions in the above-mentioned five text QA datasets, 
For the documents, to avoid the enormous cost of collecting spoken versions of long text documents, we search for off-the-shelf spoken documents relevant to each question as paired documents from the Spoken Wikipedia dataset
$^4$
% \footnote{\scriptsize{CC BY-SA 4.0 license}}
~\citep{KHN16.518}, which includes 1.2k spoken Wikipedia articles from about 400 different real speakers. %instead of using the text documents from the five text QA datasets and then collecting their speech data. 
We split the articles in Spoken Wikipedia into about 37k spoken documents with duration of 40 seconds.
%and overlap time of 10 seconds. 

% To find suitable documents for each question, 
We adopt a similar procedure with~\citet{joshi2017triviaqa} to search for relevant documents to the questions with their transcripts automatically. 
The detailed search criteria and the final number of SLUE-SQA-5 questions from each source text QA dataset are in Section~\ref{ssec:document-search} and Table~\ref{tab:num-questions} in the Appendix.

To ensure the evaluation quality, we also asked human annotators to pick 408 question-document pairs, in which the document provides enough clues to answer the question, from test data as the verified-test set.
 %Table~\ref{tab:qa-data-statistics} shows the data statistics of SLUE-SQA-5.
The data statistics of SLUE-SQA-5 are in Table~\ref{tab:qa-data-statistics}. 
% removed: ", we calculate inner products between question embeddings and document embeddings as relevance scores"

\begin{table}[!htp]\centering
\caption{SLUE-SQA-5 data statistics} \label{tab:qa-data-statistics}
\resizebox{7.5cm}{!}{%
\begin{tabular}{l|r|r|r|r}\toprule
& questions & documents & duration (h) & question speakers \\\midrule
fine-tune & 46,186 & 15,148 & 244 & 931\\
dev & 1,939 & 1,624 & 21.2 & 41\\
test & 2,382 & 1,969 & 25.8 & 51 \\
verified-test & 408 & 322 & 4.2 &  51 \\  
\bottomrule
\end{tabular}
}
\end{table}
\vspace{-.2cm}
\subsubsection{SLUE-TED for SUMM}

Of the existing corpora for abstractive speech summarization, How-2 has been used in recent work ~\citep{sharma2022end}. However, raw audio is not publicly available for the entire corpus, and the task of summarization is relatively easy due to shorter videos and simple reference summaries. Therefore, we consider the more challenging task of generating abstracts and titles for TED Talks, whose audio is publicly available. The TEDSummary dataset was introduced by ~\cite{attention-fusion} and accompanied by a tool to crawl and download TED talk videos from the web \footnote{\scriptsize {https://github.com/nttcslab-sp-admin/TEDSummary}} that may be used to recreate the TEDSummary corpus. However, the lack of information about the exact talks used in the corpus makes it difficult to reproduce their data selection. Based on their crawler, and more recent talks released on the TED website\footnote{\scriptsize {CC BY–NC–ND 4.0 license}}, we introduce SLUE-TED, a re-designed corpus of summaries for TED Talks spanning the years until 2022. 

% We evaluate our approach using TED corpus, a corpus of summaries for TED talks. This corpus is created by crawling TED talk videos along with the title and abstract for the given talk from TED website. The summary consists of both the abstract and the title of the talk. While the previous work ~\cite{attention-fusion}  releases a tool to crawl and download TED talk videos \footnote{\url{https://github.com/nttcslab-sp-admin/TEDSummary}} used in their TEDSummary corpus, the lack of information about the exact talks used in the corpus makes it difficult to exactly reproduce their data selections. We use their provided crawler to create a similar speech summarization corpus whose data statistics are reported in Table~\ref{tab:ted_stats}. A more detailed distribution of data statistics can be found in Figure~\ref{fig:summary-dist} in the Appendix.

% We observe that compared to popular speech summarization benchmarks like How2~\cite{How2}, our corpus consists of significantly longer audios and hence is a more challenging benchmark.
We find that, on average, nearly 66\% of words in the title and 57.4\% of words in the abstract are present in the transcript of a given audio, suggesting that ASR pre-training can be useful to improve speech summarization performance. For benchmark evaluation, we randomly split this corpus into 80\% finetune, 10\% validation, and 10\% test set as shown in Table~\ref{tab:ted_split}. A detailed description of the dataset is available in the Appendix~\ref{sec:summ-data-details}.

\begin{table}[!htp]\centering
\caption{SLUE-TED data split}\label{tab:ted_split}
\begin{tabular}{l|r|r}\toprule
& utterances & duration (h)\\\midrule
finetune & 3384 & 664\\
dev & 425 & 81\\
test & 424 & 84\\
\bottomrule
\end{tabular}
\end{table}
\subsubsection{SLUE-VoxPopuli for NEL}
\label{sec:nel-data-details}
SLUE-VoxPopuli was published with NER annotations in SLUE~\citep{shon2022slue}. We extend SLUE-VoxPopuli to NEL by adding word-level time stamps in the dev and test sets.  We use the Montreal Forced Aligner (MFA)~\citep{mcauliffe17_interspeech} to obtain word-level time stamps, using MFA's public English acoustic model~\citep{mfa_english_mfa_acoustic_2022}. MFA is a standard tool that is commonly used by the community to obtain ground-truth forced alignments. We manually verify the MFA produced entity alignments for 188 utterances (20\% of the utterances with entity tags) in dev set and conclude that the MFA output provides a reliable ground-truth. We share more details for the data annotation and verification procedure in Appendix~\ref{sec:nel-appendix-annot}. Data statistics for the SLUE-NEL data are shown in Table~\ref{tab:nel_stats}. Note that we do not publish NEL annotations for the {\it finetune} set as we focus on re-purposing NER models for NEL, which we believe is a more realistic use-case; as is also common for the speech-to-text forced alignment models, such as MFA, to be trained without ground-truth alignments.

\begin{table}[!htp]\centering
\caption{SLUE-NEL data statistics}\label{tab:nel_stats}
\resizebox{\linewidth}{!}{%
\begin{tabular}{l|r|r|r}\toprule
&utterances &duration (h) & \makecell{\# w/ entity tags\\(\# entity phrases)}   \\\midrule
dev & 1,750 & 5.0 & 943 (1857) \\
test & 1,838 & 4.9 & 1032 (1986) \\
\bottomrule
\end{tabular}
}
\end{table}

\section{Experiments and results} \label{sec:results}
In the SLUE Phase-1 baseline experiments, larger pre-trained models and LM shallow fusion consistently gave better performance compared to smaller pre-trained models and without LM shallow fusion. Thus, in this paper, we analyze how the ASR word error rate (WER) in pipeline models is correlated with SLU task performance, by using multiple off-the-shelf open-source ASR models, specifically NeMo models~\cite{kuchaiev2019nemo} and Whisper~\cite{radford2022robust}. Additionally, we quantify the performance gain on WER and SLU tasks achieved by fine-tuning custom ASR models compared to using off-the-shelf ASR models.

In all experiments, we use the fine-tune set of the corresponding task to fine-tune pre-trained models, the dev set to pick the best model,
% based on the corresponding task score
and the test set to evaluate both E2E and pipeline baselines. In addition, we measure the performance of an "oracle" pipeline system that uses ground-truth transcripts instead of ASR output. Below, we use the \textit{base} sized model when there are multiple variants of the pre-trained model.
\begin{figure*}[!htp]\centering
    \begin{subfigure}[b]{0.43\textwidth}
         \includegraphics[width=\textwidth]{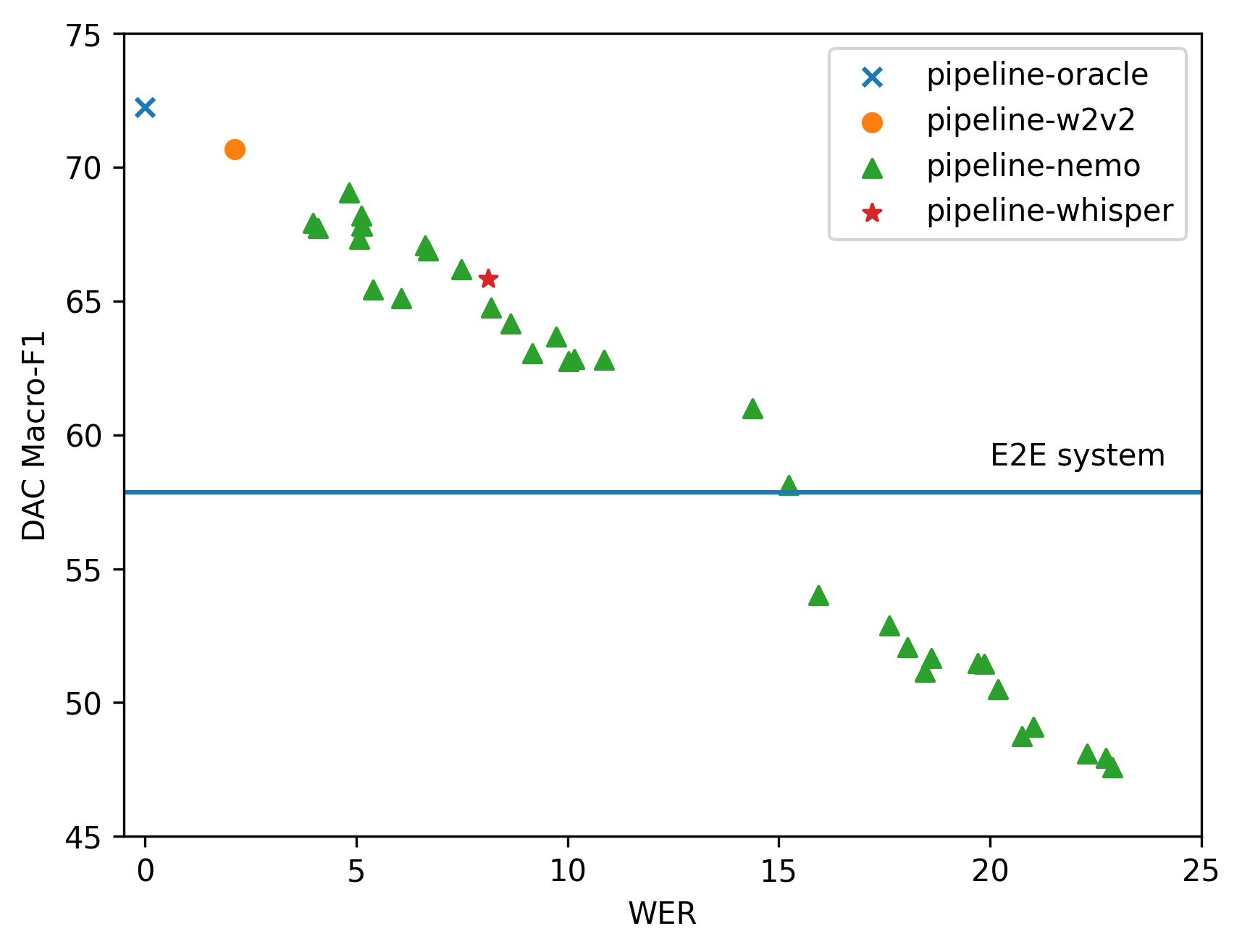}
         \vspace*{-.7cm}
         \caption{DAC task : WER and F1 score on test set}
         \vspace*{.2cm}
         \label{fig:dac_corr}
     \end{subfigure}
    % \hspace{13mm}
    \begin{subfigure}[b]{0.48\textwidth}
         \includegraphics[width=\textwidth]{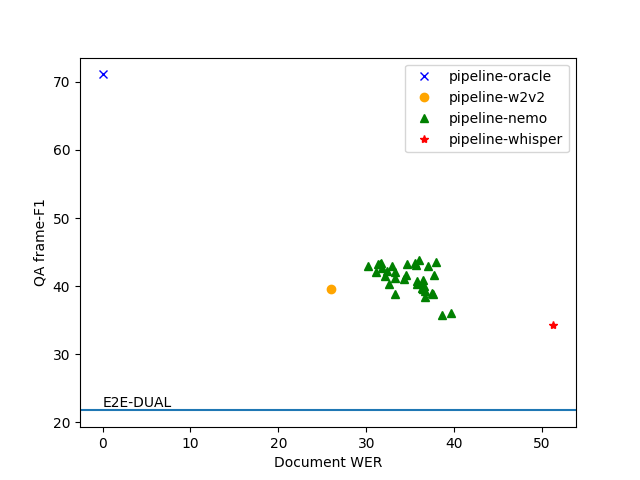}
         \vspace*{-.7cm}
         \caption{QA task: Document WER and frame-F1 scores}
         \vspace*{.2cm}         
         \label{fig:qa_corr_d}
     \end{subfigure}
     \\
     \hspace{-.5cm}
     \begin{subfigure}[b]{0.425\textwidth}
         \includegraphics[width=\textwidth]{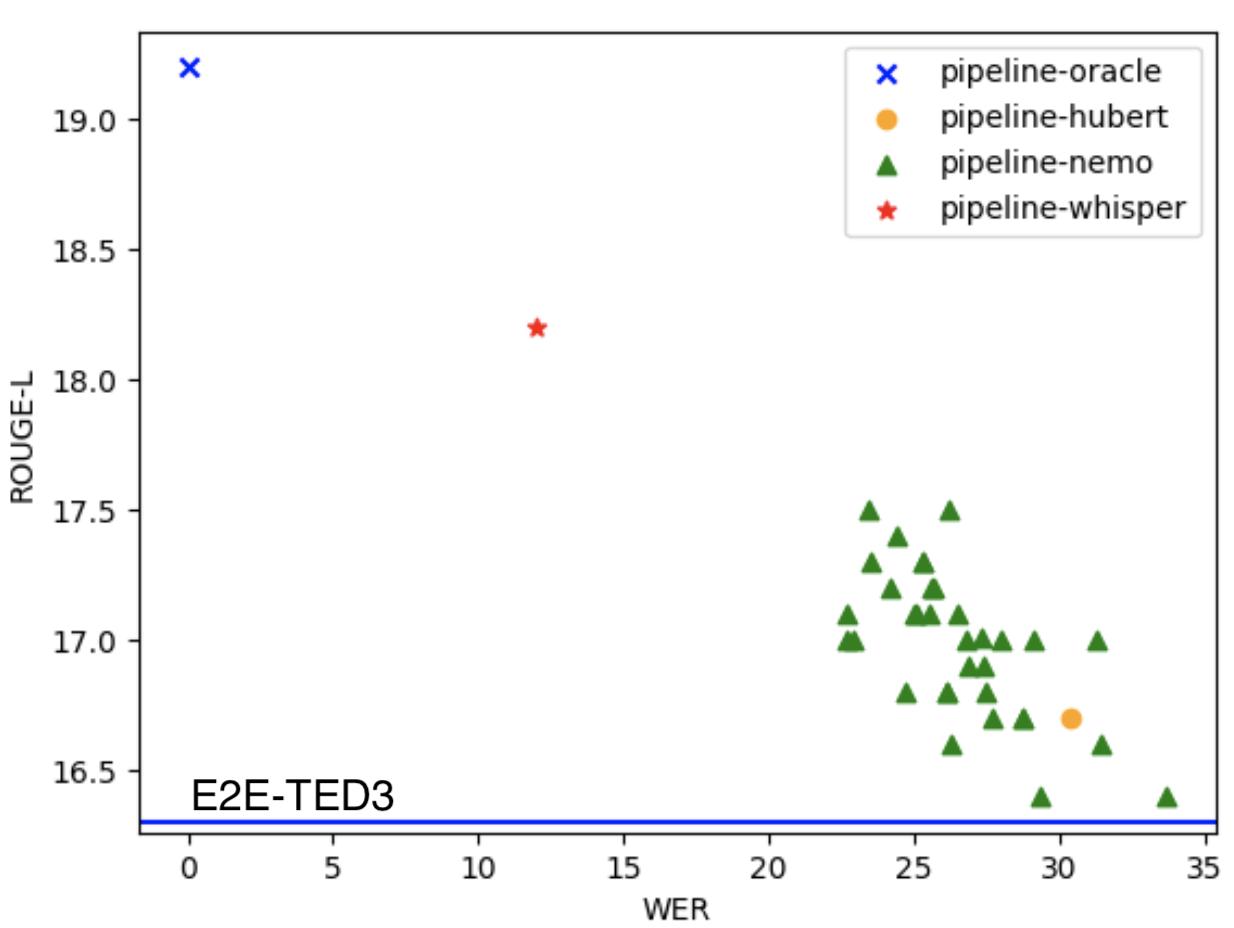}
         \vspace*{-.7cm}
         \caption{SUMM task : WER and ROUGE-L score}
         \vspace*{.2cm}         
         \label{fig:summ_corr}
     \end{subfigure}
     \hspace{5mm}
     \begin{subfigure}[b]{0.43\textwidth}
         \includegraphics[width=\textwidth]{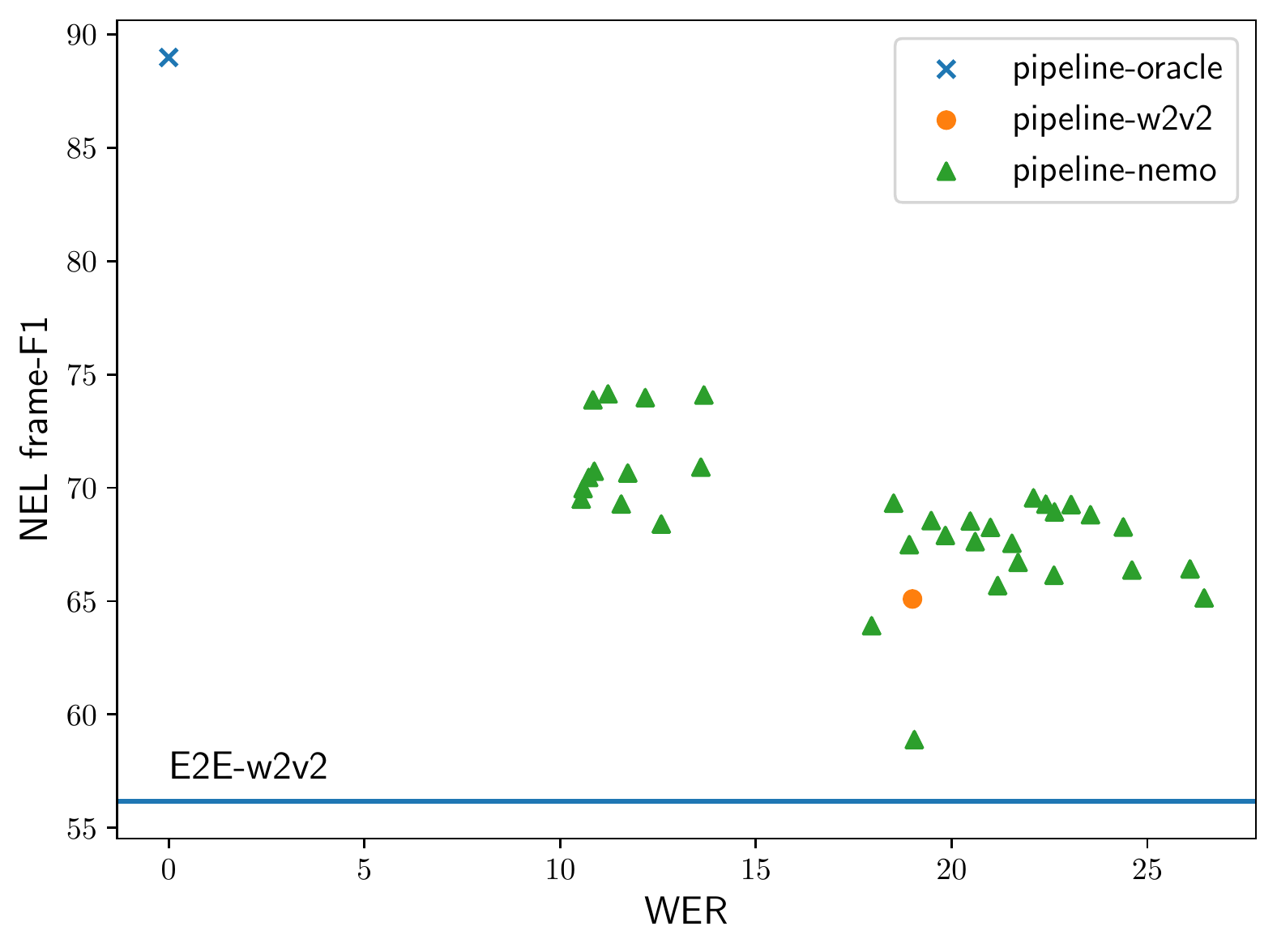}
         \vspace*{-.7cm}
         \caption{NEL task: WER and frame-F1 scores}
         \vspace*{.2cm}         
         \label{fig:nel_corr}
     \end{subfigure}
     \vspace*{-.5cm}         
    \caption{WER sensitivity on NLP model performance}
    \vspace*{-.1cm}         

    \label{fig:result_merged}
\end{figure*}

\subsection{DAC}
{\bf Baseline models}: We follow a similar setup to the sentiment analysis baseline models in SLUE Phase-1 with some differences due to the multilabel nature of DAC.  For the E2E baseline, we start with a pre-trained speech model, specifically wav2vec2~\citep{baevski2020wav2vec}, and add a self-attention pooling layer and two fully connected layers (including the output layer), with a Sigmoid output activation for each of the 18 dialog act classes.  Outputs that is higher/lower than a threshold of 0.5 are classified as positive/negative for the corresponding class.  For the pipeline baselines, we use either the off-the-shelf ASR models or an ASR using DAC data fine-tuned wav2vec2, and fine-tune a DeBERTa~\cite{he2020deberta} model for the text classification.
%The difference is that DAC is multilabel-multiclass classification. Thus, we changed cross entropy loss with Softmax to binary cross entropy loss with Sigmoid for each 18 dialog act classes. We use 0.5 as threshold to decide Sigmoid output into binary label on both validation and evaluation. We used wav2vec2-base pre-trained model to fine-tune E2E system. We use wav2vec2-base pre-train model and deberta-base model to fine-tune ASR and NLP model for pipeline system. SLUE-HVB fine-tune set is used to fine-tune the model and dev set was used to pick the best model based on macro-F1 score. We use SLUE-HVB dev and test set to evaluate E2E and pipeline baseline as shown in table~\ref{tab:dac_baseline}. Oracle pipeline system use Human transcription for input.

{\bf Results}: Table~\ref{tab:dac_baseline} shows the baseline results, and Figure~\ref{fig:dac_corr} shows the relationship between WER and F1 score of pipeline models for a variety of ASR models (the ones used in Table~\ref{tab:dac_baseline} and all other NeMo models). 
 %For correlation analysis between WER and F1 score of DAC task, we run inference on SLUE-HVB test set using available ASR models in NeMo and Whisper toolkit. The figure~\ref{fig:dac_corr} shows the evaluation results. We observed 
 We observe a strong correlation between the WER and DAC Macro F1 score (Pearson coorelation coefficient = -0.9). As the off-the-shelf ASR models perform well on conversational speech, fine-tuning the ASR model does not 
 %show a significant performance gap compared to 
 give a large improvement over the best NeMo model.

\begin{table}[!htp]\centering
\caption{DAC task baseline performance on test set. *the best NeMo model based on DAC F1 score is "stt-en-conformer-transducer-xxlarge"}\label{tab:dac_baseline}.
\resizebox{7.5cm}{!}{%
\begin{tabular}{lccc}\toprule
\multirow{2}{*}{System} &\multirow{2}{*}{\makecell{Speech\\model}} &\multirow{2}{*}{\makecell{Text\\model}} &\multicolumn{1}{c}{F1 score} \\
& & & (WER) \\\midrule
pipeline-oracle &x &DeBERTa &72.3 (0.0) \\ \midrule
pipeline-w2v2 & wav2vec2 &DeBERTa &70.7 (2.1) \\
pipeline-nemo & best model* &DeBERTa &69.1 (4.8) \\
pipeline-whisper &whisper-en &DeBERTa & 65.8 (8.1)\\ \midrule
E2E-w2v2 &wav2vec2 &x &57.9 (----) \\
\bottomrule
\end{tabular}
}
\end{table}
\vspace{-.3cm}

% \begin{figure}[ht]
% \centering
%     \includegraphics[width=0.9\linewidth]{figs/dac_wer_f1_test.png}
%     \vspace{-0.3cm}
%     \caption{DAC task : WER and F1 score on test set}
%     \label{fig:dac_corr}
%     \vspace{-0.2cm}
% \end{figure}

\subsection{QA}
% fine tune nlp model and  implementation of asr models in pipeline approach
$\textbf{Pipeline Approach:}$ The pipeline QA system is composed of an ASR model and a text QA model predicting the start and end words of the answer span on the ASR output transcript.

We fine-tuned DeBERTa with the ground-truth transcripts of the SLUE-SQA-5 fine-tune set to get the text QA model of all pipeline systems.
%We used the SLUE-SQA-5 dev set to pick the fine-tuned text QA model with the highest word-F1 score and the fine-tuned ASR model with the lowest WER for evaluation.
Note that the DeBERTa text QA models in pipeline systems and the text QA models used for searching paired documents (please refer to Section~\ref{ssec:document-search}) were fine-tuned on different datasets: the former were tuned on the SLUE-SQA-5 fine-tune set while the latter were tuned on the external SQuAD dataset. %so the question-documents pairs in the dev and test sets will not exclusively benefit our pipeline systems.
% post process prediction

% evaluation (use force alignment)
When evaluating pipeline systems on the SLUE-SQA-5 dev and test sets, we used MFA to align ground-truth transcripts and ASR output transcripts to speech.
The ground-truth answer words and the answer words predicted by the text QA model are converted to the time interval of the ground-truth and predicted answer span, which were then used to calculate the frame-F1 score.

% implementation of e2e baseline model
$\textbf{E2E Approach:}$
We used DUAL~\citep{lin2022dual} as the QA E2E approach (denoted as E2E-DUAL).
DUAL is composed of a wav2vec2-large model encoding speech waveforms, a k-means model converting wav2vec2 representations into cluster IDs, a Longformer model taking cluster IDs as input and predicting the start and end index of answer spans.
We followed the training procedure in the DUAL paper except we used the k-means model of 500 clusters and fine-tuned its Longformer model for 45 epochs on the SLUE-SQA-5 fine-tune set. 
%We picked the checkpoint with the highest frame-F1 on the dev set for evaluation.

% experimental results
$\textbf{Results:}$ 
Table~\ref{tab:qa_baseline} shows the baseline results on the test and verified-test sets, and Figure~\ref{fig:qa_corr_d} shows the relationship between document WER and frame-F1 on the test set of QA pipeline models.
We observe a strong correlation (Pearson correlation coefficient=-0.89, p-value<0.01) between document WER and frame-F1. 
Pipeline-oracle significantly outperforms all the baseline models, and the performance gap is larger in the verified-test set, suggesting that there is room for improvement. 
Besides, the pipeline-w2v2 does not outperform the pipeline-nemo model, indicating that the fine-tuned ASR model does not lead to better QA performance.

\begin{table}[!htp]\centering
\caption{QA task baseline performance. *the best Nemo model based on frame-F1 score is "stt-en-contextnet-1024".}
\label{tab:qa_baseline}
\resizebox{7.5cm}{!}{%
\begin{tabular}{lcccc}\toprule
\multirow{2}{*}{System} &\multirow{2}{*}{\makecell{Speech\\model}} &\multirow{2}{*}{\makecell{Text\\model}} &\multicolumn{2}{c}{Frame-F1 } \\\cmidrule{4-5}
& & & Test & Verified-test \\\midrule
pipeline-oracle &x & DeBERTa & 62.3 & 70.3 \\\midrule
pipeline-w2v2 & wav2vec2 & DeBERTa & 39.6  & 40.1  \\
pipeline-nemo & best model* & DeBERTa & 43.3  & 45.9  \\
pipeline-whisper &whisper-en & DeBERTa & 32.7  & 35.7 \\\midrule
E2E-DUAL & DUAL & x & 21.8 & 23.1 \\ 
\bottomrule
\end{tabular}
}
\end{table}

%%%%%%%%%%%%%%%%%%%%%%%
% old version of performance table
\iffalse
\begin{table}[!htp]\centering
\caption{QA task baseline performance. *the best Nemo model based on frame-F1 score is "stt-en-contextnet-1024". "qWER" denotes the question WER, and "dWER" denotes the document WER.}
\label{tab:qa_baseline}
\resizebox{7.5cm}{!}{%
\begin{tabular}{lccll}\toprule
\multirow{2}{*}{System} &\multirow{2}{*}{\makecell{Speech\\model}} &\multirow{2}{*}{\makecell{Text\\model}} &\multicolumn{2}{c}{Frame-F1 (qWER/ dWER)} \\\cmidrule{4-5}
& & & Test & Verified-test \\\midrule
pipeline-oracle &x & DeBERTa & 62.3 (0.0/ 0.0)& 71.1 (0.0/ 0.0) \\\midrule
pipeline-w2v2 & wav2vec2 & DeBERTa & 39.6 (19.2/ 26.0) & 40.4 (15.4/ 19.0) \\
pipeline-nemo & best model* & DeBERTa & 43.3 (8.9/ 31.5) & 45.4 (6.9/ 18.8) \\
pipeline-whisper &whisper-en & DeBERTa & 32.7 (12.4/ 51.3) & 34.3 (9.4/ 50.0)\\\midrule
E2E-DUAL & DUAL & x & 21.8 (----/ ----) & 22.0 (----/ ----) \\ 
\bottomrule
\end{tabular}
}
\end{table}
\fi
%%%%%%%%%%%%%%%%%%%%%%%%%%

% \begin{figure}[ht]
% \centering
%     \includegraphics[width=0.95\linewidth]{figs/qa_qwer_ff1.png}
%     \caption{QA task: Question WER and frame-F1 scores}
%     \label{fig:qa_corr_q}%
% \end{figure}
\subsection{SUMM}
\begin{table*}[!htp]\centering
\caption{SUMM task baseline performance. The ASR models are trained on the TEDLIUM-3 corpus. *the best NeMo model based on SUMM ROUGE-L score is "conformer-transducer-xxlarge". For
pipeline models, we also experiment with training NLU model on ASR Transcripts (ASR) instead of ground
truth transcript. }
\label{tab:summ_baseline}
\resizebox{16cm}{!}{%
\begin{tabular}{lccccccccc}\toprule
System & \makecell{Speech\\model} & \makecell{Text\\model} & ROUGE-1 & ROUGE-2 & ROUGE-L & METEOR & BERTScore & WER\\
\midrule
pipeline-oracle &x & LED &  30.1 & 7.7 & 19.3 & 13.7 & 83.8 & \hphantom{0}0.0\\ \hline 
pipeline-w2v2 & wav2vec2-ASR & LED & 26.8 & 5.1 &  16.8 & 12.4 & 82.5 & 34.4\\
pipeline-hubert & Hubert-ASR & LED & 26.9 & 5.3 & 16.7 & 12.6 & 82.5 & 30.4\\
% 25.5 &	4.5 &	16.0 & 11.7 &	82.3 \\
% pipeline-hubert & Hubert & LED (ASR) & 28.2 & 5.9 & 17.9 & 14.3 & 83.2	 \\
pipeline-nemo & best model* & LED & 27.6 & 6.2 & 17.5 & 13.0 & 82.4 & 23.4\\
pipeline-whisper & whisper-en & LED & 28.6 & 6.7 & 18.2 & 12.9 & 83.4 & 12.0\\
pipeline-whisper ASR & whisper-en & LED(ASR) &  29.0 & 7.0 & 18.6 & 13.0 & 83.7& 12.0\\ \hline
E2E-TED3 & TEDLIUM-3 Conformer & x & 23.8 &	5.1 &	16.3 &	11.7 & 84.0 & ----\\
% Pipeline-300s & LED(ASR) &  &  &  &  & \\ \hline
\bottomrule
\end{tabular}
}
\end{table*}
% Speech summarization can be performed using pipeline and end-to-end approaches. Pipeline approaches use speech recognition to transcribe utterances within TED talks followed by text summarization on the resulting transcript. 

% We use the open-source speech processing toolkit ESPnet~\cite{watanabe2018espnet} to compute baseline results for the SUMM task.

\textbf{Pipeline Approach}: The oracle pipeline is constructed by using the ground truth transcript to train a text summarization model, and infer the most likely summary from the ground truth transcript. Then, we use different combinations of speech recognizers and text summarization models to build different pipeline models for speech summarization. For the pipeline baseline, we train ASR models on the TEDLIUM-3 ~\cite{TEDLIUM-3} corpus using the ESPNet~\cite{watanabe2018espnet}  toolkit. The ASR models consist of a conformer encoder-decoder architecture with pre-trained SSL representations as features (see Appendix \ref{sec:summ-appendix-hyperparam} for more details about our models). We also experiment with state-of-the-art off-the-shelf speech recognizers, including Whisper ~\cite{radford2022robust} and NeMo models.  The resulting talk transcripts are very long, often exceeding 2048 tokens, requiring our text summarization models to be able to handle such long input sequences. Therefore, we use the Longformer Encoder-Decoder (LED-large) model ~\cite{beltagy2020longformer}, initialized using BART-large model~\cite{BART}. We investigate training our text summarisation model on both ground truth and ASR transcripts.

\textbf{E2E Approach}: E2E speech summarization model is trained using the ESPNet ~\cite{watanabe2018espnet} toolkit by first pre-training for speech recognition task on the TEDLIUM-3 corpus ~\cite{TEDLIUM-3} and then fine-tuning on our SLUE-TED data for speech summarization task as described in ~\cite{sharma2022end}. 

% We use the open-source speech processing toolkit ESPnet~\cite{watanabe2018espnet} to compute baseline results for the SUMM task. We first compute text summarization results by fine-tuning LED~\cite{Longformer}, which is a pretrained LM based on Longformer~\cite{Longformer} encoder-decoder architecture that can process long text sequences. For pipeline experiments, our ASR model is a conformer encoder-decoder architecture that uses pre-trained Hubert~\cite{hsu2021hubert} representations as features. Our ASR model is trained on TEDLIUM-3~\cite{TEDLIUM-3} corpus\footnote{We use TEDLIUM-3 corpus to train our ASR model since they are too few audios in our training set to train a robust ASR model}.
% Since the SLUE-TED corpus consists of significantly long audios, we use Voice Activity Detection\footnote{\url{https://chromium.googlesource.com/external/webrtc/+/master/common_audio/vad/}, Agressiveness=1} to segment the speech input into smaller utterances and then compute the inference of our ASR model on these utterances.
% For the NLU model, we experiment with training on both ground truth transcripts and transcripts produced by our ASR model.  
% Inspired by strong results from prior work~\cite{sharma2022end} on E2E Speech summarization, we train a baseline E2E model, which is initialized using our ASR model. Since the audios were too long to fit the entire speech input on a GPU, we trained our E2E model using only the first 30,000 speech frames. 

\textbf{Results:} Table~\ref{tab:summ_baseline} shows the performance for all baseline models on the test set (see Appendix \ref{sec:summ-res-details} for dev set performance). We observe that the performance of the pipeline system can be improved by using a strong ASR model like Whisper. Further, we observe that the pipeline system performs slightly better when the text summarization model is fine-tuned on ASR transcripts. The pipeline models outperform the E2E system on ROUGE and METEOR, showing that the pipeline model aids in producing more accurate words. However, the end-to-end model does have a higher BERTScore, demonstrating the ability of the E2E model to produce semantically relevant summaries. All the baseline models perform worse than the pipeline-oracle model suggesting room for improvement.

To analyze the correlation between WER and the performance of the speech summarization task, we plot ROUGE-L scores in Figure~\ref{fig:summ_corr} for various pipeline systems and a ground-truth transcript-based text summarization model. 
% For Nemo conformer and squeezeformer models, the audio is too long to perform inference using a GPU, and hence we have to break audio input into 5-minute chunks and perform inference separately on each of these chunks. The ATS model for all these systems is trained using ground truth transcripts. 
We observe a strong correlation (Pearson correlation coefficient=-0.9, p-value$<$0.01) between WER and ROUGE-L scores, suggesting that we can boost SUMM performance using a stronger ASR model.
% Correlation=(-0.9063803478704322, 2.8178268110668017e-14)

% \begin{figure}[ht]
% \centering
%     \includegraphics[width=0.9\linewidth]{figs/summ_WER_rouge_final.png}
%     \caption{SUMM task : WER and ROUGE-L score}
%     \label{fig:summ_corr}
% \end{figure}

To facilitate a better understanding of the performance of our E2E SUMM model, we analyze the percentage of exact matches in reference summary and predicted summaries for each POS tag.
% The POS tags are computed automatically using SPACY\footnote{\url{https://spacy.io/}} toolkit. 
We observe that the majority of summarization errors occur because the model is not able to correctly generate the proper nouns in summary. A similar analysis on the percentage of exact matches for named entities shows that only 6.6\% of entities in the reference summary were found in the predicted summary. Based on this analysis, we infer that the current speech summarization models struggle to correctly extract entities for the summary. (Full POS tags match available in Table~\ref{tab:exact:matches} in the Appendix)

\subsection{NEL}
{\bf Baseline models}:
For NEL inference, we use the baseline NER models from \citet{shon2022slue}. 
Both the E2E and ASR (within pipeline) models use wav2vec2 as the backbone and are trained with character-level connectionist temporal classification (CTC)~\citep{graves2006connectionist}.
% Both the E2E model and the ASR component in the pipeline NER model are character-level connectionist temporal classification (CTC)~\citep{graves2006connectionist} models that use the wav2vec2.0 Base pre-trained speech model as the backbone.
%and are trained with character-level connectionist temporal classification (CTC) loss~\citep{graves2006connectionist}. 
The text NER (within pipeline) model uses the DeBERTa as the backbone and is trained on ground-truth transcripts. 
% The text NER component of the pipeline model uses the DeBERTa-Base~\citep{he2020deberta} pre-trained model as the backbone and is trained on ground-truth transcripts. 
%The E2E model %training, the transcript is modified with 
%is trained to output entity-specific special character tokens as delimiters around the entity phrases. 
Note that no dedicated model is trained for NEL.  This is intentional: NER and NEL are related tasks and a realistic use case would require a single model that performs both tasks. 

% \begin{figure}[ht]
%     \includegraphics[width=\linewidth]{figs/nel_ppl.pdf}
%     \vspace{-0.7cm}
%     \caption{Example inference for pipeline NEL model.}
%     \label{fig:nel_ppl}
%     \vspace{-0.5cm}
% \end{figure}

\begin{figure}[ht]
    \includegraphics[width=\linewidth]{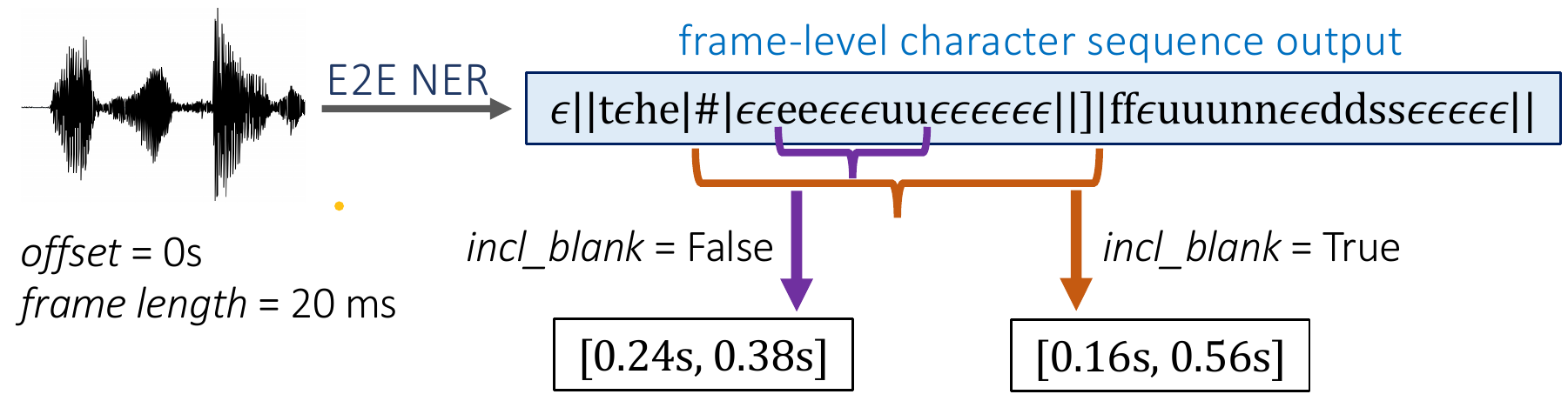}
    \vspace{-0.5cm}
    \caption{Example inference for an E2E NEL model using a CTC recognizer. The transcript is ``the eu funds''. `\#' and `]' are the start and end labels of an ORG entity.}
    \label{fig:nel_e2e}
    \vspace{-0.3cm}
\end{figure}

{\bf Inference}:
A CTC model produces a posterior probability matrix, $\mathcal{E} \in \mathbb{R}^{T\times V}$, consisting of the posterior of each character in the vocabulary of size $V$ for each of the $T$ frames in the input audio.  For ASR, the character vocabulary consists of the English alphabet, a word separator token ``|'', and a blank token``$\epsilon$''. For the E2E model, the vocabulary also includes special characters for the start and end of an entity phrase.
%We adopt the greedy decoding strategy on the 
We obtain a frame-level character sequence output via greedy decoding on $\mathcal{E}$.
%the CTC model's posterior probability matrix, $\mathcal{E} \in \mathbb{R}^{T\times V}$, where $T$ is the number of input frames. 
%This gives frame-level character prediction. 
The time stamps corresponding to ``|'' tokens in the output character sequence provide word-level start and end boundaries.  
% Next, we de-duplicate the resulting character sequence and convert frame indices to time using the stride value. This gives a character sequence and the corresponding time sequence denoting character onsets. The duration from the onset of the current character to the onset of the following character is regarded as the character-level time stamp. The contiguous character-level time stamps between two-word separator tokens are merged to obtain word-level time stamps. Since the onset time stamps obtained from CTC are known to be only approximate estimators\textcolor{red}{ToDo: Add reference},
As CTC is not trained with an explicit alignment signal, the word boundary tokens may not be a reliable indicator of the true time stamps, and we introduce two hyperparameters as a heuristic fix for possible mis-alignments: {\it offset} is a fixed duration 
%variable that 
by which we shift the time stamp predictions, and
%by a fixed value before evaluation. 
{\it incl\_blank} $\in \{0,1\}$ 
%is a Boolean value that 
denotes whether any trailing $\epsilon$ tokens are considered a part of the predicted entity segment.
%time stamp. 
%An example is presented in Figure~\ref{fig:nel_e2e}.
% When {\it incl\_blank} is False, the predicted word-level segments are shorter.  

In the pipeline approach, the predicted text from ASR is passed to a text NER model, and the time stamps for detected entities are extracted from the ASR's $\mathcal{E}$. For the E2E model, the time stamps corresponding to the entity start and end special characters are extracted directly from its $\mathcal{E}$. An example is presented in Fig.~\ref{fig:nel_e2e}.

\begin{table}[!htp]\centering
\caption{NEL task baseline performance on test set. The W2V2-B models are fine-tuned on slue-voxpopuli data.*the best nemo model based on NEL frame-f1 score on dev is ``stt\_en\_conformer\_ctc\_small"}
\label{tab:nel_baseline}
\resizebox{7.5cm}{!}{%
\begin{tabular}{lcc|c|c}\toprule
\multirow{2}{*}{System} &\multirow{2}{*}{\makecell{Speech\\model}} & \multirow{2}{*}{\makecell{Text\\model}} & \multirow{2}{*}{frame-F1} & \multicolumn{1}{c}{word-F1} \\
& & &  & ($\rho$=0.8)  \\\midrule
pipeline-oracle & x & DeBERTa & 89.0 & 90.0  \\
pipeline-w2v2 & wav2vec2 & DeBERTa & 65.2 & 72.0  \\
E2E-w2v2 & wav2vec2 & x & 56.3 & 59.6  \\
pipeline-nemo & best model* & DeBERTa & 74.1 & 81.4 \\
\bottomrule
\end{tabular}
}
\end{table}
{\bf Results}: Table~\ref{tab:nel_baseline} presents the baseline results.
%{\it frame-F1} and {\it word-F1} results. 
The pipeline and E2E baselines have %strikingly 
fairly similar frame-F1, but these approaches have complementary strengths as seen from their precision and recall values (see Table~\ref{tab:nel_prec_recall_analysis}, Appendix \ref{sec:nel-appendix-error}). 
% The E2E model outperforms in {\it precision} (i.e, more predicted regions are named entities), whereas the pipeline model outperforms in {\it recall}. 
% The mismatch in text NER's training (ground-truth text) and inference (ASR output) could lead to higher false positives in the pipeline model. 
% , which could be due to the train-test mismatch in the text NER. 
% For word-F1, relaxing the tolerance from $\rho=1$ to $\rho=0.8$ gives a major performance boost -- up to 30\% and 116\% relative for pipeline and E2E models respectively.
% higher tolerance (lower $\rho$) improves the score as expected.  We also find that the off-the-shelf NeMo ASR model 
%from NeMo toolkit 
We also find that the off-the-shelf NeMo ASR model
({\it pipeline-nemo}) outperforms the dataset-specific ASR model ({\it pipeline-w2v2}).\footnote{More word-F1 results in Tab.~\ref{tab:nel_baseline_all} in Appendix~\ref{sec:nel-appendix-additional}.}
%has better performance, and relaxing the strict matching criteria ($\rho=1$) to a tolerance of 80\% overlap ($\rho=0.8$) gives a significant improvement. 
% As NLP topline has access to the ground truth transcript, the alignments are the same as the ground truth time stamps obtained from forced alignment. So the only errors are the mistakes made by the text NER model, and thus the {\it word-F1} measure is not affected by $\rho$.

Figure~\ref{fig:nel_corr} shows a scatter plot of NEL and WER scores for a variety of pipeline models.
% using the off-the-shelf ASR models from the NeMo toolkit.
 Although models with the lowest WER do have the best frame-F1, the overall correlation is not high.
 %scores are not well-correlated. 
 % Interestingly, the same does not hold for NER, where F1 and WER are well-correlated (see Figure~\ref{fig:ner_corr}, Appendix~\ref{sec:nel-appendix-error}).
 The NeMo models have different training objectives and model architectures, and we note that within each model class, the ASR and NEL metrics are much better correlated (see Figure~\ref{fig:nel_nemo_corr}, Appendix~\ref{sec:nel-appendix-error}). This suggests that model architecture and/or training objective play a significant role in alignment quality.\footnote{The details of hyperparameter tuning and timestamp extraction from NeMo models are in Appendix~\ref{sec:nel-appendix-hyperparam}.}

% \begin{figure}[ht]
% \centering
%     \includegraphics[width=0.9\linewidth]{figs/nel_test_wer_f1_best_params.pdf}
%     \vspace{-0.2cm}
%     \caption{NEL task: Test set WER and frame-F1 scores.}
%     \label{fig:nel_corr}
%     \vspace{-0.5cm}
% \end{figure}

\section{Discussion}
Among the baseline models, our pipeline models generally outperform their end-to-end counterparts. However, as shown in prior work (e.g.,~\citep{prosody-useful, Pasad2021OnTU}), end-to-end models often have more room for improvement with careful and creative modeling ideas, and we hope that this new testbed helps spur such research.  

In addition, the WER sensitivity analysis in Figure~\ref{fig:result_merged} suggests different strategies are needed for the pipeline system depending on the SLU task. For example, fine-tuned ASR (pipeline-w2v2) plays a significant role in the DAC task while the QA task is not, and ASR model architecture is critical for the NEL task while WER is more matter for DAC and SUMM tasks.

\section{Conclusion} \label{sec:conclusion}
% SLUE Phase-1 introduced a new benchmark and two SLU tasks (sentiment analysis and NER). 
SLUE Phase-2, with four additional SLU tasks and high-quality annotation, enables a more comprehensive analysis of diverse SLU tasks than previously possible.  Besides the task definitions and annotations, this work contributes multiple baselines and performance analysis using modern off-the-shelf ASR and text models.  
% Among the baseline models, our pipeline models generally outperform their end-to-end counterparts.  However, as shown in prior work (e.g.,~\citep{prosody-useful, Pasad2021OnTU}), end-to-end models often have more room for improvement with careful and creative modeling ideas, and we hope that this new testbed helps spur such research.  
The baseline performance on all tasks is far from perfect, and the relative performance of different models differs across tasks, indicating that these tasks are ripe for additional work and analysis to push the boundary of SLU research.  
%In the future, we will maintain the SLUE benchmark toolkit and closely work with the potential contributors in this benchmark to advance the boundary of SLU-related research. Additional tasks will also be released in the SLUE benchmark in the future.

% \section*{Limitations}
% ACL 2023 requires all submissions to have a section titled ``Limitations'', for discussing the limitations of the paper as a complement to the discussion of strengths in the main text. This section should occur after the conclusion, but before the references. It will not count towards the page limit.
% The discussion of limitations is mandatory. Papers without a limitation section will be desk-rejected without review.

% While we are open to different types of limitations, just mentioning that a set of results have been shown for English only probably does not reflect what we expect. 
% Mentioning that the method works mostly for languages with limited morphology, like English, is a much better alternative.
% In addition, limitations such as low scalability to long text, the requirement of large GPU resources, or other things that inspire crucial further investigation are welcome.

% \section*{Ethics Statement}
% Scientific work published at ACL 2023 must comply with the ACL Ethics Policy.\footnote{\url{https://www.aclweb.org/portal/content/acl-code-ethics}} We encourage all authors to include an explicit ethics statement on the broader impact of the work, or other ethical considerations after the conclusion but before the references. The ethics statement will not count toward the page limit (8 pages for long, 4 pages for short papers).

\section*{Limitations}
One limitation of this work is the lack of human performance scores on the new tasks.  Although the baseline performance is far from perfect, and it seems quite likely that human performance is much better, this should be measured in future work.  Another limitation is that it is unknown how much each task should benefit from access to the audio in addition to text; this could be measured in principle for humans, but again we leave this to future work.

% Though the tasks chosen in this paper are important and practically relevant, the set of tasks thus chosen is not all-encompassing, and future work may expand this selection with more tasks. 

\section*{Broader Impact and Ethics}
%In this paper, we have proposed an extension of the SLUE benchmark for Spoken Language Understanding tasks by describing new dataset releases for Spoken Dialog Act Classification, Spoken Question Answering, Speech Summarization, and Spoken Named Entity Localization.  
Spoken language understanding benchmarks, like the ones we propose in this work, 
%The release of new benchmarks will propel the development of spoken language understanding and intelligent spoken interfaces. Our work may be 
facilitate the development of technologies that may be particularly useful for speakers who are unable to read or write text and ultimately also
%may serve as inspiration to set up similar benchmarks 
for unwritten languages, where speech is the only form of communication.  We hope that this work also spurs more collaboration across the fields of speech and natural language processing, both of which are needed to make progress in this area.  
%The initial end-to-end systems presented in this paper, though they do not outperform the best pipeline in most cases, demonstrate how speech attributes that go beyond transcripts can be used in these tasks, setting the stage for future work on end-to-end models. 

%The spoken language understanding systems whose development may be aided by our work may find application in a variety of fields- including customer service help desks, call centers, meeting rooms, lecture halls, search and retrieval engines, and conversational agents. 

% Our work utilizes speech from the user in order to predict attributes- whereas the pipeline models use only the transcript, end-to-end models may use other attributes of the speaker. Privacy-preserving methods, particularly for end-to-end models may be necessary to 

We ensured that the SLUE-SQA speech data collection from AMT was conducted with a higher wage (on average, US\$10 per hour) than the US federal minimum wage. This wage includes compensation for the time spent on re-recording and addressing technical issues on the recording platform. We further took measures to ensure that our data collection and annotation process did not introduce any potential biases in the SLUE Phase-2 benchmark.
Specifically, for SLUE-SQA, we implemented an automatic check using the Google Speech-to-Text service. If the Word Error Rate (WER) exceeded 30\%, workers were recommended to re-record the utterance. We chose a 30\% WER threshold to identify and exclude empty or prematurely cut utterances. Our analysis showed that such violations were less than 8\% of questions. Additionally, we personally listened to each recording and only discarded those where a significant portion of the content was missing. Recordings were accepted even if the WER exceeded 30\%, ensuring that our dataset does not include any potential bias inherent in the automated speech-to-text service.

The DAC annotation in SLUE-HVB and verified-test set in SLUE-SQA data were done by ASAPP internal data labeling team. Everyone who participated in the annotation was an employee of ASAPP and conducted the work within the scope of their usual employment. Specifically, most of them have over 1 year of experience in speech and language-related data labeling and their education level is above a Master's degree.

\section*{Acknowledgements}
We would like to thank Kyle Hager, and Molly Ruhl for their helpful comments and discussion from a linguistic perspective, and the whole ASAPP MLDL team members for high quality annotation. Part of this work used PSC Bridges2 and NCSA Delta through allocations CIS210014 and IRI120015 from the Advanced Cyberinfrastructure Coordination Ecosystem: Services \& Support (ACCESS) program,
which is supported by National Science Foundation grants
\#2138259, \#2138286, \#2138307, \#2137603, and \#2138296.

% This document has been adapted by Jordan Boyd-Graber, Naoaki Okazaki, Anna Rogers from the style files used for earlier ACL, EMNLP and NAACL proceedings, including those for
% EACL 2023 by Isabelle Augenstein and Andreas Vlachos,
% EMNLP 2022 by Yue Zhang, Ryan Cotterell and Lea Frermann,
% ACL 2020 by Steven Bethard, Ryan Cotterell and Rui Yan,
% ACL 2019 by Douwe Kiela and Ivan Vuli\'{c},
% NAACL 2019 by Stephanie Lukin and Alla Roskovskaya, 
% ACL 2018 by Shay Cohen, Kevin Gimpel, and Wei Lu, 
% NAACL 2018 by Margaret Mitchell and Stephanie Lukin,
% Bib\TeX{} suggestions for (NA)ACL 2017/2018 from Jason Eisner,
% ACL 2017 by Dan Gildea and Min-Yen Kan, NAACL 2017 by Margaret Mitchell, 
% ACL 2012 by Maggie Li and Michael White, 
% ACL 2010 by Jing-Shin Chang and Philipp Koehn, 
% ACL 2008 by Johanna D. Moore, Simone Teufel, James Allan, and Sadaoki Furui, 
% ACL 2005 by Hwee Tou Ng and Kemal Oflazer, 
% ACL 2002 by Eugene Charniak and Dekang Lin, 
% and earlier ACL and EACL formats written by several people, including
% John Chen, Henry S. Thompson and Donald Walker.
% Additional elements were taken from the formatting instructions of the \emph{International Joint Conference on Artificial Intelligence} and the \emph{Conference on Computer Vision and Pattern Recognition}.

\bibliography{custom}

\begin{thebibliography}{74}
\expandafter\ifx\csname natexlab\endcsname\relax\def\natexlab#1{#1}\fi

\bibitem[{Arora et~al.(2022{\natexlab{a}})Arora, Dalmia, Chang, Yan, Black, and
  Watanabe}]{prosody-useful}
Siddhant Arora, Siddharth Dalmia, Xuankai Chang, Brian Yan, Alan~W. Black, and
  Shinji Watanabe. 2022{\natexlab{a}}.
\newblock \href {https://doi.org/10.21437/Interspeech.2022-10890} {Two-pass low
  latency end-to-end spoken language understanding}.
\newblock In \emph{Interspeech}.

\bibitem[{Arora et~al.(2022{\natexlab{b}})Arora, Dalmia, Denisov, Chang, Ueda,
  Peng, Zhang, Kumar, Ganesan, Yan et~al.}]{arora2022espnet}
Siddhant Arora, Siddharth Dalmia, Pavel Denisov, Xuankai Chang, Yushi Ueda,
  Yifan Peng, Yuekai Zhang, Sujay Kumar, Karthik Ganesan, Brian Yan, et~al.
  2022{\natexlab{b}}.
\newblock {ESPnet-SLU}: Advancing spoken language understanding through
  {ESP}net.
\newblock In \emph{ICASSP}.

\bibitem[{Baevski et~al.(2022)Baevski, Hsu, Xu, Babu, Gu, and
  Auli}]{Baevski2022data2vecAG}
Alexei Baevski, Wei-Ning Hsu, Qiantong Xu, Arun Babu, Jiatao Gu, and Michael
  Auli. 2022.
\newblock data2vec: A general framework for self-supervised learning in speech,
  vision and language.
\newblock In \emph{International Conference on Machine Learning}.

\bibitem[{Baevski et~al.(2020)Baevski, Zhou, Mohamed, and
  Auli}]{baevski2020wav2vec}
Alexei Baevski, Yuhao Zhou, Abdelrahman Mohamed, and Michael Auli. 2020.
\newblock wav2vec 2.0: A framework for self-supervised learning of speech
  representations.
\newblock \emph{NeurIPS}.

\bibitem[{Baril et~al.(2022)Baril, Cardinal, and Koerich}]{baril2022named}
Guillaume Baril, Patrick Cardinal, and Alessandro~Lameiras Koerich. 2022.
\newblock Named entity recognition for audio de-identification.
\newblock \emph{arXiv preprint arXiv:2204.12622}.

\bibitem[{Bastianelli et~al.(2020)Bastianelli, Vanzo, Swietojanski, and
  Rieser}]{bastianelli2020slurp}
Emanuele Bastianelli, Andrea Vanzo, Pawel Swietojanski, and Verena Rieser.
  2020.
\newblock {SLURP}: A spoken language understanding resource package.
\newblock In \emph{EMNLP}.

\bibitem[{Baudi{\v{s}} and {\v{S}}ediv{\`y}(2015)}]{baudivs2015modeling}
Petr Baudi{\v{s}} and Jan {\v{S}}ediv{\`y}. 2015.
\newblock Modeling of the question answering task in the {YodaQA} system.
\newblock In \emph{International Conference of the Cross-Language Evaluation
  Forum for European Languages}, pages 222--228.

\bibitem[{Beltagy et~al.(2020)Beltagy, Peters, and
  Cohan}]{beltagy2020longformer}
Iz~Beltagy, Matthew~E Peters, and Arman Cohan. 2020.
\newblock Longformer: The long-document transformer.
\newblock \emph{arXiv preprint arXiv:2004.05150}.

\bibitem[{Berant et~al.(2013)Berant, Chou, Frostig, and
  Liang}]{berant2013semantic}
Jonathan Berant, Andrew Chou, Roy Frostig, and Percy Liang. 2013.
\newblock Semantic parsing on {F}reebase from question-answer pairs.
\newblock In \emph{EMNLP}.

\bibitem[{Boersma and Weenink(2009)}]{Boersma2009}
Paul Boersma and David Weenink. 2009.
\newblock \href {http://www.praat.org} {Praat: doing phonetics by computer
  (version 5.1.13)}.

\bibitem[{Busso et~al.(2008)Busso, Bulut, Lee, Kazemzadeh, Mower, Kim, Chang,
  Lee, and Narayanan}]{busso2008iemocap}
Carlos Busso, Murtaza Bulut, Chi-Chun Lee, Abe Kazemzadeh, Emily Mower, Samuel
  Kim, Jeannette~N Chang, Sungbok Lee, and Shrikanth~S Narayanan. 2008.
\newblock {IEMOCAP}: Interactive emotional dyadic motion capture database.
\newblock In \emph{Language resources and evaluation}.

\bibitem[{Calhoun et~al.(2010)Calhoun, Carletta, Brenier, Mayo, Jurafsky,
  Steedman, and Beaver}]{calhoun2010nxt}
Sasha Calhoun, Jean Carletta, Jason~M Brenier, Neil Mayo, Dan Jurafsky, Mark
  Steedman, and David Beaver. 2010.
\newblock The {NXT}-format {S}witchboard corpus: a rich resource for
  investigating the syntax, semantics, pragmatics and prosody of dialogue.
\newblock \emph{Language resources and evaluation}.

\bibitem[{Chen et~al.(2020{\natexlab{a}})Chen, Lu, Xu, Cao, Zhang, and
  Fan}]{chen2020large}
Eric~Y. Chen, Zhiyun Lu, Hao Xu, Liangliang Cao, Yu~Zhang, and James Fan.
  2020{\natexlab{a}}.
\newblock {A large scale speech sentiment corpus}.
\newblock In \emph{Language resources and evaluation}.

\bibitem[{Chen et~al.(2021)Chen, Wang, Chen, Wu, Liu, Chen, Li, Kanda,
  Yoshioka, Xiao, Wu, Zhou, Ren, Qian, Qian, Zeng, and Wei}]{Chen2021WavLMLS}
Sanyuan Chen, Chengyi Wang, Zhengyang Chen, Yu~Wu, Shujie Liu, Zhuo Chen, Jinyu
  Li, Naoyuki Kanda, Takuya Yoshioka, Xiong Xiao, Jian Wu, Long Zhou, Shuo Ren,
  Yanmin Qian, Yao Qian, Micheal Zeng, and Furu Wei. 2021.
\newblock Wav{LM}: Large-scale self-supervised pre-training for full stack
  speech processing.
\newblock \emph{IEEE Journal of Selected Topics in Signal Processing},
  16:1505--1518.

\bibitem[{Chen et~al.(2020{\natexlab{b}})Chen, Ita~Levitan, Levine, Mandic, and
  Hirschberg}]{chen2020acoustic}
Xi~Leslie Chen, Sarah Ita~Levitan, Michelle Levine, Marko Mandic, and Julia
  Hirschberg. 2020{\natexlab{b}}.
\newblock Acoustic-prosodic and lexical cues to deception and trust:
  deciphering how people detect lies.
\newblock \emph{Transactions of the Association for Computational Linguistics},
  8:199--214.

\bibitem[{Chuang et~al.(2020)Chuang, Liu, Lee, and Lee}]{chuang2020speechbert}
Yung-Sung Chuang, Chi-Liang Liu, Hung-Yi Lee, and Lin-shan Lee. 2020.
\newblock Speech{BERT}: An audio-and-text jointly learned language model for
  end-to-end spoken question answering.
\newblock In \emph{Interspeech}.

\bibitem[{Cohn et~al.(2019)Cohn, Laish, Beryozkin, Li, Shafran, Szpektor,
  Hartman, Hassidim, and Matias}]{cohn2019audio}
Ido Cohn, Itay Laish, Genady Beryozkin, Gang Li, Izhak Shafran, Idan Szpektor,
  Tzvika Hartman, Avinatan Hassidim, and Yossi Matias. 2019.
\newblock \href {https://doi.org/10.18653/v1/n19-2025} {{Audio
  de-identification: A new entity recognition task}}.
\newblock In \emph{NAACL}.

\bibitem[{Coucke et~al.(2018)Coucke, Saade, Ball, Bluche, Caulier, Leroy,
  Doumouro, Gisselbrecht, Caltagirone, Lavril et~al.}]{coucke2018snips}
Alice Coucke, Alaa Saade, Adrien Ball, Th{\'e}odore Bluche, Alexandre Caulier,
  David Leroy, Cl{\'e}ment Doumouro, Thibault Gisselbrecht, Francesco
  Caltagirone, Thibaut Lavril, et~al. 2018.
\newblock Snips voice platform: an embedded spoken language understanding
  system for private-by-design voice interfaces.
\newblock \emph{arXiv:1805.10190}.

\bibitem[{Denkowski and Lavie(2014)}]{denkowski-lavie-2014-meteor}
Michael Denkowski and Alon Lavie. 2014.
\newblock \href {https://doi.org/10.3115/v1/W14-3348} {Meteor universal:
  Language specific translation evaluation for any target language}.
\newblock In \emph{Proceedings of the Ninth Workshop on Statistical Machine
  Translation}, pages 376--380, Baltimore, Maryland, USA. Association for
  Computational Linguistics.

\bibitem[{Evain et~al.(2021)Evain, Nguyen, Le, Boito, Mdhaffar, Alisamir, Tong,
  Tomashenko, Dinarelli, Parcollet et~al.}]{evain2021lebenchmark}
Sol{\`e}ne Evain, Ha~Nguyen, Hang Le, Marcely~Zanon Boito, Salima Mdhaffar,
  Sina Alisamir, Ziyi Tong, Natalia Tomashenko, Marco Dinarelli, Titouan
  Parcollet, et~al. 2021.
\newblock Le{B}enchmark: A reproducible framework for assessing self-supervised
  representation learning from speech.
\newblock In \emph{Interspeech}.

\bibitem[{Feng et~al.(2021)Feng, Yu, Cai, Liu, Zheng, and
  Wang}]{feng2021asrglue}
Lingyun Feng, Jianwei Yu, Deng Cai, Songxiang Liu, Haitao Zheng, and Yan Wang.
  2021.
\newblock \href {http://arxiv.org/abs/2108.13048} {{ASR-GLUE: A New Multi-task
  Benchmark for ASR-Robust Natural Language Understanding}}.
\newblock \emph{arXiv:2108.13048}.

\bibitem[{Graves et~al.(2006)Graves, Fern{\'a}ndez, Gomez, and
  Schmidhuber}]{graves2006connectionist}
Alex Graves, Santiago Fern{\'a}ndez, Faustino Gomez, and J{\"u}rgen
  Schmidhuber. 2006.
\newblock Connectionist temporal classification: labelling unsegmented sequence
  data with recurrent neural networks.
\newblock In \emph{International Conference on Machine Learning}.

\bibitem[{He et~al.(2020)He, Liu, Gao, and Chen}]{he2020deberta}
Pengcheng He, Xiaodong Liu, Jianfeng Gao, and Weizhu Chen. 2020.
\newblock De{BERT}a: Decoding-enhanced {BERT} with disentangled attention.
\newblock In \emph{ICLR}.

\bibitem[{Hemphill et~al.(1990)Hemphill, Godfrey, and
  Doddington}]{hemphill1990atis}
Charles~T. Hemphill, John~J. Godfrey, and George~R. Doddington. 1990.
\newblock \href {https://doi.org/10.3115/116580.116613} {{The ATIS spoken
  language systems pilot corpus}}.
\newblock In \emph{Speech and Natural Language}.

\bibitem[{Hernandez et~al.(2018)Hernandez, Nguyen, Ghannay, Tomashenko, and
  Est{\`{e}}ve}]{TEDLIUM-3}
Fran{\c{c} }ois Hernandez, Vincent Nguyen, Sahar Ghannay, Natalia Tomashenko,
  and Yannick Est{\`{e}}ve. 2018.
\newblock \href {https://doi.org/10.1007/978-3-319-99579-3_21} {{TED}-{LIUM} 3:
  Twice as much data and corpus repartition for experiments on speaker
  adaptation}.
\newblock In \emph{Speech and Computer}, pages 198--208. Springer International
  Publishing.

\bibitem[{Hsu et~al.(2021)Hsu, Bolte, Tsai, Lakhotia, Salakhutdinov, and
  Mohamed}]{hsu2021hubert}
Wei-Ning Hsu, Benjamin Bolte, Yao-Hung~Hubert Tsai, Kushal Lakhotia, Ruslan
  Salakhutdinov, and Abdelrahman Mohamed. 2021.
\newblock \href {http://arxiv.org/abs/2106.07447} {{HuBERT: Self-Supervised
  Speech Representation Learning by Masked Prediction of Hidden Units}}.
\newblock \emph{arXiv:2106.07447}.

\bibitem[{Janin et~al.(2003)Janin, Baron, Edwards, Ellis, Gelbart, Morgan,
  Peskin, Pfau, Shriberg, Stolcke, and Wooters}]{janin2003icsi}
A.~Janin, D.~Baron, J.~Edwards, D.~Ellis, D.~Gelbart, N.~Morgan, B.~Peskin,
  T.~Pfau, E.~Shriberg, A.~Stolcke, and C.~Wooters. 2003.
\newblock \href {https://doi.org/10.1109/ICASSP.2003.1198793} {The icsi meeting
  corpus}.
\newblock In \emph{2003 IEEE International Conference on Acoustics, Speech, and
  Signal Processing, 2003. Proceedings. (ICASSP '03).}, volume~1, pages I--I.

\bibitem[{Joshi et~al.(2017)Joshi, Choi, Weld, and
  Zettlemoyer}]{joshi2017triviaqa}
Mandar Joshi, Eunsol Choi, Daniel~S Weld, and Luke Zettlemoyer. 2017.
\newblock Trivia{QA}: A large scale distantly supervised challenge dataset for
  reading comprehension.
\newblock In \emph{ACL}.

\bibitem[{Jurafsky et~al.(1998)Jurafsky, Shriberg, Fox, and
  Curl}]{jurafsky1998lexical}
Dan Jurafsky, Elizabeth Shriberg, Barbara Fox, and Traci Curl. 1998.
\newblock Lexical, prosodic, and syntactic cues for dialog acts.
\newblock In \emph{Discourse Relations and Discourse Markers}.

\bibitem[{Kano et~al.(2021)Kano, Ogawa, Delcroix, and
  Watanabe}]{attention-fusion}
Takatomo Kano, Atsunori Ogawa, Marc Delcroix, and Shinji Watanabe. 2021.
\newblock Attention-based multi-hypothesis fusion for speech summarization.
\newblock In \emph{IEEE Automatic Speech Recognition and Understanding Workshop
  (ASRU)}.

\bibitem[{K{\"o}hn et~al.(2016)K{\"o}hn, Stegen, and Baumann}]{KHN16.518}
Arne K{\"o}hn, Florian Stegen, and Timo Baumann. 2016.
\newblock Mining the spoken {W}ikipedia for speech data and beyond.
\newblock In \emph{Language Resources and Evaluation}, Paris, France. European
  Language Resources Association (ELRA).

\bibitem[{Kuchaiev et~al.(2019)Kuchaiev, Li, Nguyen, Hrinchuk, Leary, Ginsburg,
  Kriman, Beliaev, Lavrukhin, Cook et~al.}]{kuchaiev2019nemo}
Oleksii Kuchaiev, Jason Li, Huyen Nguyen, Oleksii Hrinchuk, Ryan Leary, Boris
  Ginsburg, Samuel Kriman, Stanislav Beliaev, Vitaly Lavrukhin, Jack Cook,
  et~al. 2019.
\newblock Ne{M}o: a toolkit for building {AI} applications using neural
  modules.
\newblock \emph{arXiv preprint arXiv:1909.09577}.

\bibitem[{Kwiatkowski et~al.(2019)Kwiatkowski, Palomaki, Redfield, Collins,
  Parikh, Alberti, Epstein, Polosukhin, Devlin, Lee
  et~al.}]{kwiatkowski2019natural}
Tom Kwiatkowski, Jennimaria Palomaki, Olivia Redfield, Michael Collins, Ankur
  Parikh, Chris Alberti, Danielle Epstein, Illia Polosukhin, Jacob Devlin,
  Kenton Lee, et~al. 2019.
\newblock Natural questions: a benchmark for question answering research.
\newblock \emph{Transactions of the Association for Computational Linguistics},
  7:453--466.

\bibitem[{Lai et~al.(2020)Lai, Chuang, yi~Lee, Li, and
  Glass}]{Lai2020SemiSupervisedSL}
Cheng-I Lai, Yung-Sung Chuang, Hung yi~Lee, Shang-Wen Li, and James~R. Glass.
  2020.
\newblock Semi-supervised spoken language understanding via self-supervised
  speech and language model pretraining.
\newblock \emph{ICASSP}.

\bibitem[{Lee et~al.(2018)Lee, Wu, Liu, and Lee}]{lee2018spoken}
Chia-Hsuan Lee, Szu-Lin Wu, Chi-Liang Liu, and Hung-yi Lee. 2018.
\newblock Spoken {SQuAD}: A study of mitigating the impact of speech
  recognition errors on listening comprehension.
\newblock \emph{Interspeech}, pages 3459--3463.

\bibitem[{Lewis et~al.(2019)Lewis, Liu, Goyal, Ghazvininejad, Mohamed, Levy,
  Stoyanov, and Zettlemoyer}]{BART}
Mike Lewis, Yinhan Liu, Naman Goyal, Marjan Ghazvininejad, Abdelrahman Mohamed,
  Omer Levy, Ves Stoyanov, and Luke Zettlemoyer. 2019.
\newblock \href {https://doi.org/10.48550/ARXIV.1910.13461} {Bart: Denoising
  sequence-to-sequence pre-training for natural language generation,
  translation, and comprehension}.

\bibitem[{Lin(2004)}]{lin-2004-rouge}
Chin-Yew Lin. 2004.
\newblock \href {https://aclanthology.org/W04-1013} {{ROUGE}: A package for
  automatic evaluation of summaries}.
\newblock In \emph{Text Summarization Branches Out}, pages 74--81, Barcelona,
  Spain. Association for Computational Linguistics.

\bibitem[{Lin et~al.(2022{\natexlab{a}})Lin, Chuang, Chung, Yang, Chen, Li,
  Mohamed, Lee, and Lee}]{lin2022dual}
Guan-Ting Lin, Yung-Sung Chuang, Ho-Lam Chung, Shu-wen Yang, Hsuan-Jui Chen,
  Shang-Wen Li, Abdelrahman Mohamed, Hung-yi Lee, and Lin-shan Lee.
  2022{\natexlab{a}}.
\newblock Dual: Textless spoken question answering with speech discrete unit
  adaptive learning.
\newblock \emph{arXiv preprint arXiv:2203.04911}.

\bibitem[{Lin et~al.(2022{\natexlab{b}})Lin, Lee, and Tang}]{lin2022melhubert}
Tzu-Quan Lin, Hung-yi Lee, and Hao Tang. 2022{\natexlab{b}}.
\newblock Melhubert: A simplified hubert on mel spectrogram.
\newblock \emph{arXiv preprint arXiv:2211.09944}.

\bibitem[{Liu et~al.(2019)Liu, Eshghi, Swietojanski, and
  Rieser}]{liu2019benchmarking}
Xingkun Liu, Arash Eshghi, Pawel Swietojanski, and Verena Rieser. 2019.
\newblock Benchmarking natural language understanding services for building
  conversational agents.
\newblock \emph{arXiv preprint arXiv:1903.05566}.

\bibitem[{Lugosch et~al.(2021{\natexlab{a}})Lugosch, Papreja, Ravanelli, Heba,
  and Parcollet}]{lugosch2021timers}
Loren Lugosch, Piyush Papreja, Mirco Ravanelli, Abdelwahab Heba, and Titouan
  Parcollet. 2021{\natexlab{a}}.
\newblock Timers and such: A practical benchmark for spoken language
  understanding with numbers.
\newblock \emph{arXiv preprint arXiv:2104.01604}.

\bibitem[{Lugosch et~al.(2021{\natexlab{b}})Lugosch, Papreja, Ravanelli, Heba,
  and Parcollet}]{Lugosch2021TimersAS}
Loren Lugosch, Piyush Papreja, Mirco Ravanelli, Abdelwahab Heba, and Titouan
  Parcollet. 2021{\natexlab{b}}.
\newblock Timers and such: A practical benchmark for spoken language
  understanding with numbers.
\newblock \emph{ArXiv}, abs/2104.01604.

\bibitem[{Lugosch et~al.(2019)Lugosch, Ravanelli, Ignoto, Tomar, and
  Bengio}]{lugosch2019speech}
Loren Lugosch, Mirco Ravanelli, Patrick Ignoto, Vikrant~Singh Tomar, and Yoshua
  Bengio. 2019.
\newblock \href {https://doi.org/10.21437/Interspeech.2019-2396} {{Speech model
  pre-training for end-to-end spoken language understanding}}.
\newblock In \emph{INTERSPEECH}.

\bibitem[{Martinez-Lucas et~al.(2020)Martinez-Lucas, Abdelwahab, and
  Busso}]{martinez2020msp}
Luz Martinez-Lucas, Mohammed Abdelwahab, and Carlos Busso. 2020.
\newblock The {MSP-C}onversation corpus.
\newblock In \emph{INTERSPEECH}.

\bibitem[{McAuliffe et~al.(2017)McAuliffe, Socolof, Mihuc, Wagner, and
  Sonderegger}]{mcauliffe17_interspeech}
Michael McAuliffe, Michaela Socolof, Sarah Mihuc, Michael Wagner, and Morgan
  Sonderegger. 2017.
\newblock \href {https://doi.org/10.21437/Interspeech.2017-1386} {{Montreal
  Forced Aligner: Trainable Text-Speech Alignment Using Kaldi}}.
\newblock In \emph{Interspeech}, pages 498--502.

\bibitem[{McAuliffe and Sonderegger(2022)}]{mfa_english_mfa_acoustic_2022}
Michael McAuliffe and Morgan Sonderegger. 2022.
\newblock English mfa acoustic model v2.0.0.
\newblock Technical report,
  \url{https://mfa-models.readthedocs.io/acoustic/English/English MFA acoustic
  model v2_0_0.html}.

\bibitem[{McCowan et~al.(2005)McCowan, Carletta, Kraaij, Ashby, Bourban, Flynn,
  Guillemot, Hain, Kadlec, Karaiskos, Kronenthal, Lathoud, Lincoln, Lisowska,
  Post, Reidsma, and Wellner}]{mccowan2005ami}
I.~McCowan, J.~Carletta, W.~Kraaij, S.~Ashby, S.~Bourban, M.~Flynn,
  M.~Guillemot, T.~Hain, J.~Kadlec, V.~Karaiskos, M.~Kronenthal, G.~Lathoud,
  M.~Lincoln, A.~Lisowska, W.~Post, Dennis Reidsma, and P.~Wellner. 2005.
\newblock The ami meeting corpus.
\newblock In \emph{Proceedings of Measuring Behavior 2005, 5th International
  Conference on Methods and Techniques in Behavioral Research}, pages 137--140.
  Noldus Information Technology.

\bibitem[{Mohamed et~al.(2022)Mohamed, Lee, Borgholt, Havtorn, Edin, Igel,
  Kirchhoff, Li, Livescu, Maaløe, Sainath, and Watanabe}]{mohamed2022self}
Abdelrahman Mohamed, Hung-yi Lee, Lasse Borgholt, Jakob~D. Havtorn, Joakim
  Edin, Christian Igel, Katrin Kirchhoff, Shang-Wen Li, Karen Livescu, Lars
  Maaløe, Tara~N. Sainath, and Shinji Watanabe. 2022.
\newblock Self-supervised speech representation learning: A review.
\newblock \emph{IEEE Journal of Selected Topics in Signal Processing},
  16(6):1179--1210.

\bibitem[{Panayotov et~al.(2015)Panayotov, Chen, Povey, and
  Khudanpur}]{panayotov2015librispeech}
Vassil Panayotov, Guoguo Chen, Daniel Povey, and Sanjeev Khudanpur. 2015.
\newblock \href {https://doi.org/10.1109/ICASSP.2015.7178964} {{Librispeech: An
  ASR corpus based on public domain audio books}}.
\newblock In \emph{ICASSP}.

\bibitem[{Pasad et~al.(2021)Pasad, Wu, Shon, Livescu, and Han}]{Pasad2021OnTU}
Ankita Pasad, Felix Wu, Suwon Shon, Karen Livescu, and Kyu~J. Han. 2021.
\newblock On the use of external data for spoken named entity recognition.
\newblock In \emph{North American Chapter of the Association for Computational
  Linguistics}.

\bibitem[{Peng et~al.(2022)Peng, Arora, Higuchi, Ueda, Kumar, Ganesan, Dalmia,
  Chang, and Watanabe}]{pretrain-E2E-SLU}
Yifan Peng, Siddhant Arora, Yosuke Higuchi, Yushi Ueda, Sujay Kumar, Karthik
  Ganesan, Siddharth Dalmia, Xuankai Chang, and Shinji Watanabe. 2022.
\newblock A study on the integration of pre-trained ssl, asr, lm and slu models
  for spoken language understanding.
\newblock \emph{arXiv preprint arXiv:2211.05869}.

\bibitem[{Radford et~al.(2022)Radford, Kim, Xu, Brockman, McLeavey, and
  Sutskever}]{radford2022robust}
Alec Radford, Jong~Wook Kim, Tao Xu, Greg Brockman, Christine McLeavey, and
  Ilya Sutskever. 2022.
\newblock Robust speech recognition via large-scale weak supervision.
\newblock \emph{arXiv preprint arXiv:2212.04356}.

\bibitem[{Rajpurkar et~al.(2016)Rajpurkar, Zhang, Lopyrev, and
  Liang}]{Rajpurkar2016SQuAD1Q}
Pranav Rajpurkar, Jian Zhang, Konstantin Lopyrev, and Percy Liang. 2016.
\newblock Squad: 100,000+ questions for machine comprehension of text.
\newblock In \emph{Conference on Empirical Methods in Natural Language
  Processing}.

\bibitem[{Reddy et~al.(2019)Reddy, Chen, and Manning}]{reddy2019coqa}
Siva Reddy, Danqi Chen, and Christopher~D Manning. 2019.
\newblock Coqa: A conversational question answering challenge.
\newblock \emph{Transactions of the Association for Computational Linguistics},
  7:249--266.

\bibitem[{Reimers and Gurevych(2019)}]{reimers2019sentence}
Nils Reimers and Iryna Gurevych. 2019.
\newblock Sentence-bert: Sentence embeddings using siamese bert-networks.
\newblock In \emph{Proceedings of the 2019 Conference on Empirical Methods in
  Natural Language Processing and the 9th International Joint Conference on
  Natural Language Processing (EMNLP-IJCNLP)}, pages 3982--3992.

\bibitem[{Robertson et~al.(2009)Robertson, Zaragoza
  et~al.}]{robertson2009probabilistic}
Stephen Robertson, Hugo Zaragoza, et~al. 2009.
\newblock The probabilistic relevance framework: Bm25 and beyond.
\newblock \emph{Foundations and Trends{\textregistered} in Information
  Retrieval}, 3(4):333--389.

\bibitem[{Sanabria et~al.(2018)Sanabria, Caglayan, Palaskar, Elliott, Barrault,
  Specia, and Metze}]{sanabria2018how2}
Ramon Sanabria, Ozan Caglayan, Shruti Palaskar, Desmond Elliott, Lo{\"\i}c
  Barrault, Lucia Specia, and Florian Metze. 2018.
\newblock \href {https://arxiv.org/abs/1811.00347} {How2: a large-scale dataset
  for multimodal language understanding}.
\newblock \emph{arXiv preprint arXiv:1811.00347}.

\bibitem[{Sharma et~al.(2022)Sharma, Palaskar, Black, and
  Metze}]{sharma2022end}
Roshan Sharma, Shruti Palaskar, Alan~W Black, and Florian Metze. 2022.
\newblock End-to-end speech summarization using restricted self-attention.
\newblock In \emph{ICASSP 2022-2022 IEEE International Conference on Acoustics,
  Speech and Signal Processing (ICASSP)}, pages 8072--8076. IEEE.

\bibitem[{Shon et~al.(2022{\natexlab{a}})Shon, Pasad, Wu, Brusco, Artzi,
  Livescu, and Han}]{shon2022slue}
Suwon Shon, Ankita Pasad, Felix Wu, Pablo Brusco, Yoav Artzi, Karen Livescu,
  and Kyu~J Han. 2022{\natexlab{a}}.
\newblock Slue: New benchmark tasks for spoken language understanding
  evaluation on natural speech.
\newblock In \emph{ICASSP 2022-2022 IEEE International Conference on Acoustics,
  Speech and Signal Processing (ICASSP)}, pages 7927--7931. IEEE.

\bibitem[{Shon et~al.(2022{\natexlab{b}})Shon, Wu, Kim, Sridhar, Livescu, and
  Watanabe}]{Shon2022ContextawareFO}
Suwon Shon, Felix Wu, Kwangyoun Kim, Prashant Sridhar, Karen Livescu, and
  Shinji Watanabe. 2022{\natexlab{b}}.
\newblock Context-aware fine-tuning of self-supervised speech models.
\newblock \emph{arXiv preprint arXiv:2212.08542}.

\bibitem[{Tomasello et~al.(2022)Tomasello, Shrivastava, Lazar, Hsu, Le, Sagar,
  Elkahky, Copet, Hsu, Mordechay et~al.}]{tomasello2022stop}
Paden Tomasello, Akshat Shrivastava, Daniel Lazar, Po-Chun Hsu, Duc Le, Adithya
  Sagar, Ali Elkahky, Jade Copet, Wei-Ning Hsu, Yossef Mordechay, et~al. 2022.
\newblock Stop: A dataset for spoken task oriented semantic parsing.
\newblock \emph{arXiv preprint arXiv:2207.10643}.

\bibitem[{Tomashenko et~al.(2019)Tomashenko, Caubri{\`e}re, Est{\`e}ve,
  Laurent, and Morin}]{tomashenko2019recent}
Natalia Tomashenko, Antoine Caubri{\`e}re, Yannick Est{\`e}ve, Antoine Laurent,
  and Emmanuel Morin. 2019.
\newblock Recent advances in end-to-end spoken language understanding.
\newblock In \emph{7th International Conference on Statistical Language and
  Speech Processing (SLSP)}.

\bibitem[{Tran et~al.(2018)Tran, Toshniwal, Bansal, Gimpel, Livescu, and
  Ostendorf}]{tran2018parsing}
Trang Tran, Shubham Toshniwal, Mohit Bansal, Kevin Gimpel, Karen Livescu, and
  Mari Ostendorf. 2018.
\newblock Parsing speech: A neural approach to integrating lexical and
  acoustic-prosodic information.
\newblock In \emph{Proceedings of NAACL-HLT}.

\bibitem[{Wang et~al.(2018)Wang, Singh, Michael, Hill, Levy, and
  Bowman}]{wang2018glue}
Alex Wang, Amanpreet Singh, Julian Michael, Felix Hill, Omer Levy, and Samuel~R
  Bowman. 2018.
\newblock {GLUE}: A multi-task benchmark and analysis platform for natural
  language understanding.
\newblock In \emph{ICLR}.

\bibitem[{Wang et~al.(2021)Wang, Riviere, Lee, Wu, Talnikar, Haziza,
  Williamson, Pino, and Dupoux}]{wang2021voxpopuli}
Changhan Wang, Morgane Riviere, Ann Lee, Anne Wu, Chaitanya Talnikar, Daniel
  Haziza, Mary Williamson, Juan Pino, and Emmanuel Dupoux. 2021.
\newblock \href {https://doi.org/10.18653/v1/2021.acl-long.80} {{VoxPopuli: A
  Large-Scale Multilingual Speech Corpus for Representation Learning,
  Semi-Supervised Learning and Interpretation}}.
\newblock \emph{arXiv:2101.00390}.

\bibitem[{Watanabe et~al.(2018)Watanabe, Hori, Karita, Hayashi, Nishitoba,
  Unno, Soplin, Heymann, Wiesner, Chen, Renduchintala, and
  Ochiai}]{watanabe2018espnet}
Shinji Watanabe, Takaaki Hori, Shigeki Karita, Tomoki Hayashi, Jiro Nishitoba,
  Yuya Unno, Nelson Enrique~Yalta Soplin, Jahn Heymann, Matthew Wiesner, Nanxin
  Chen, Adithya Renduchintala, and Tsubasa Ochiai. 2018.
\newblock \href {https://doi.org/10.21437/Interspeech.2018-1456} {{ESPNet:
  End-to-end speech processing toolkit}}.
\newblock In \emph{INTERSPEECH}.

\bibitem[{Wu et~al.(2022{\natexlab{a}})Wu, Kim, Pan, Han, Weinberger, and
  Artzi}]{wu2022performance}
Felix Wu, Kwangyoun Kim, Jing Pan, Kyu~J Han, Kilian~Q Weinberger, and Yoav
  Artzi. 2022{\natexlab{a}}.
\newblock Performance-efficiency trade-offs in unsupervised pre-training for
  speech recognition.
\newblock In \emph{ICASSP 2022-2022 IEEE International Conference on Acoustics,
  Speech and Signal Processing (ICASSP)}, pages 7667--7671. IEEE.

\bibitem[{Wu et~al.(2022{\natexlab{b}})Wu, Kim, Watanabe, Han, McDonald,
  Weinberger, and Artzi}]{wu2022wav2seq}
Felix Wu, Kwangyoun Kim, Shinji Watanabe, Kyu Han, Ryan McDonald, Kilian~Q
  Weinberger, and Yoav Artzi. 2022{\natexlab{b}}.
\newblock Wav2seq: Pre-training speech-to-text encoder-decoder models using
  pseudo languages.
\newblock \emph{arXiv preprint arXiv:2205.01086}.

\bibitem[{Wu et~al.(2020)Wu, Nafziger, Scodary, and
  Maas}]{wu2020harpervalleybank}
Mike Wu, Jonathan Nafziger, Anthony Scodary, and Andrew Maas. 2020.
\newblock Harpervalleybank: A domain-specific spoken dialog corpus.
\newblock \emph{arXiv preprint arXiv:2010.13929}.

\bibitem[{Yadav et~al.(2020)Yadav, Ghosh, Yu, and Shah}]{yadav2020end}
Hemant Yadav, Sreyan Ghosh, Yi~Yu, and Rajiv~Ratn Shah. 2020.
\newblock \href {https://doi.org/10.21437/Interspeech.2020-2482} {{End-to-end
  named entity recognition from English speech}}.
\newblock In \emph{INTERSPEECH}.

\bibitem[{Yang et~al.(2021)Yang, Chi, Chuang, Lai, Lakhotia, Lin, Liu, Shi,
  Chang, Lin et~al.}]{yang2021superb}
Shu-wen Yang, Po-Han Chi, Yung-Sung Chuang, Cheng-I~Jeff Lai, Kushal Lakhotia,
  Yist~Y Lin, Andy~T Liu, Jiatong Shi, Xuankai Chang, Guan-Ting Lin, et~al.
  2021.
\newblock {SUPERB}: Speech processing universal performance benchmark.
\newblock In \emph{INTERSPEECH}.

\bibitem[{You et~al.(2022)You, Chen, Liu, Ge, Wu, and Zou}]{you2022end}
Chenyu You, Nuo Chen, Fenglin Liu, Shen Ge, Xian Wu, and Yuexian Zou. 2022.
\newblock End-to-end spoken conversational question answering: Task, dataset
  and model.
\newblock \emph{arXiv preprint arXiv:2204.14272}.

\bibitem[{Zadeh et~al.(2018)Zadeh, Liang, Vanbriesen, Poria, Tong, Cambria,
  Chen, and Morency}]{zadeh2018multimodal}
Amir Zadeh, Paul~Pu Liang, Jonathan Vanbriesen, Soujanya Poria, Edmund Tong,
  Erik Cambria, Minghai Chen, and Louis~Philippe Morency. 2018.
\newblock \href {https://doi.org/10.18653/v1/p18-1208} {{Multimodal language
  analysis in the wild: CMU-MOSEI dataset and interpretable dynamic fusion
  graph}}.
\newblock In \emph{ACL}.

\bibitem[{Zhang* et~al.(2020)Zhang*, Kishore*, Wu*, Weinberger, and
  Artzi}]{bert-score}
Tianyi Zhang*, Varsha Kishore*, Felix Wu*, Kilian~Q. Weinberger, and Yoav
  Artzi. 2020.
\newblock \href {https://openreview.net/forum?id=SkeHuCVFDr} {Bertscore:
  Evaluating text generation with bert}.
\newblock In \emph{International Conference on Learning Representations}.

\end{thebibliography}
\bibliographystyle{acl_natbib}
\newpage
\onecolumn
\appendix

\noindent\textbf{Appendix}
% \section{DAC}
% \subsection{Dialog act list}
% \label{sec:dac_appendix}

% \input{tabs/da_detail.tex}
% \begin{figure}[ht]
%     \centering
%     \includegraphics[width=0.8\linewidth]{figs/dac_wer_f1_dev.png}
%     \caption{DAC task: WER and F1 scores on dev set}
%     \label{fig:dac_corr_dev}
% \end{figure}

\section{DAC}
\subsection{Dialog act list}
\label{sec:dac_appendix}

Figure~\ref{fig:dac_corr_dev} shows the corelation between WER and F1 score on dev set. Table~\ref{tab:dac_baseline_full} shows the experiment result including dev set.
%Please add the following packages if necessary:
%\usepackage{booktabs, multirow} % for borders and merged ranges
%\usepackage{soul}% for underlines
%\usepackage[table]{xcolor} % for cell colors
%\usepackage{changepage,threeparttable} % for wide tables
%If the table is too wide, replace \begin{table}[!htp]...\end{table} with
%\begin{adjustwidth}{-2.5 cm}{-2.5 cm}\centering\begin{threeparttable}[!htb]...\end{threeparttable}\end{adjustwidth}
\begin{table}[!htp]\centering
\caption{Dialog acts detail}\label{tab:da_detail}
\scriptsize
\resizebox{16cm}{!}{%
\begin{tabular}{p{0.09\linewidth} | p{0.15\linewidth} | p{0.3\linewidth} | p{0.15\linewidth}}
\toprule
\textbf{actions} &\textbf{sub-actions} &\textbf{Definition} &\textbf{example} \\\midrule
\multirow{5}{*}{question} &\textbf{question\_check} &Questions that check/verify information unique to a listener &What is your address? \\ \cline{2-4}
&\textbf{question\_repeat} &Requests for someone to repeat what they said in order to clarify/understand &Can you repeat that please? \\ \cline{2-4}
&\textbf{question\_general} &All other questions &How can I help you today? \\ \midrule
\multirow{4}{*}{answer} &\textbf{answer\_agree} &Answers indicating a positive response or acceptance &Yeah, let’s do that \\ \cline{2-4}
&\textbf{answer\_dis} &Answers indicating a negative response or denial &No, that’s okay \\\cline{2-4}
&\textbf{answer\_general} &All other answers & \\ \midrule
\multirow{21}{*}{statement} &\textbf{apology} &A number of often-templated utterances indicating a speaker is apologetic &I’m sorry to hear that! \\ \cline{2-4}
&\textbf{thanks} &A number of often-templated utterances indicating a speaker is appreciative &Thanks for doing that \\ \cline{2-4}
&\textbf{acknowledge} &A response indicating that a speaker has heard, or is empathizing with, what another speaker has said &Ok / I understand \\ \cline{2-4}
&\textbf{statement\_open} &Formulaic opening statements that might contain a greeting, introduction, or some other pleasantries &Hi my name is XX \\ \cline{2-4}
&\textbf{statement\_close} &Formulaic closing statements indicating that the conversation is coming to an end, often containing salutations &Have a great day \\ \cline{2-4}
&\textbf{statement\_problem} &An utterance that contains a user’s primary reason for calling in (this may include questions if the question clearly indicates the call reason) &I lost my debit card / I just called in because I wanted to know what are my local branch hours? \\ \cline{2-4}
&\textbf{statement\_instruct} &An imperative utterance that indicates the speaker wants the listener to do something &Go to the website and log in / You’ll need to upload a copy of your form \\ \cline{2-4}
&\textbf{statement\_general} &All other statements & \\ \midrule
\multirow{10}{*}{natural speech} &\textbf{backchannel} &Verbal or non-verbal expressions indicating the listener’s attention, agreement, or understanding, while not having much significant meaning on their own &uh-huh / is that right? \\ \cline{2-4}
&\textbf{disfluency} &filler, reparandum, interregnum &Uh../ uh no… / debit uh no (credit card) \\ \cline{2-4}
&\textbf{self} &Essentially rhetorical utterances, or utterances where a speaker is not expecting a response from the listener (i.e. talking to one’s self) &Oh, look at me I’ve forgotten which button to press here / Hmm now where did I put that other number… \\ \midrule
other &\textbf{other} &Any utterances that don’t fit in any of the above categories, including noise, gibberish, or otherwise uninterpretable speech &[noise] / fjdskl / ///////// \\ 
\bottomrule
\end{tabular}
}
\end{table}

\clearpage
\subsection{Annotation detail}
\begin{figure}[htp!]
    \centering
    \includegraphics[width=0.8\linewidth]{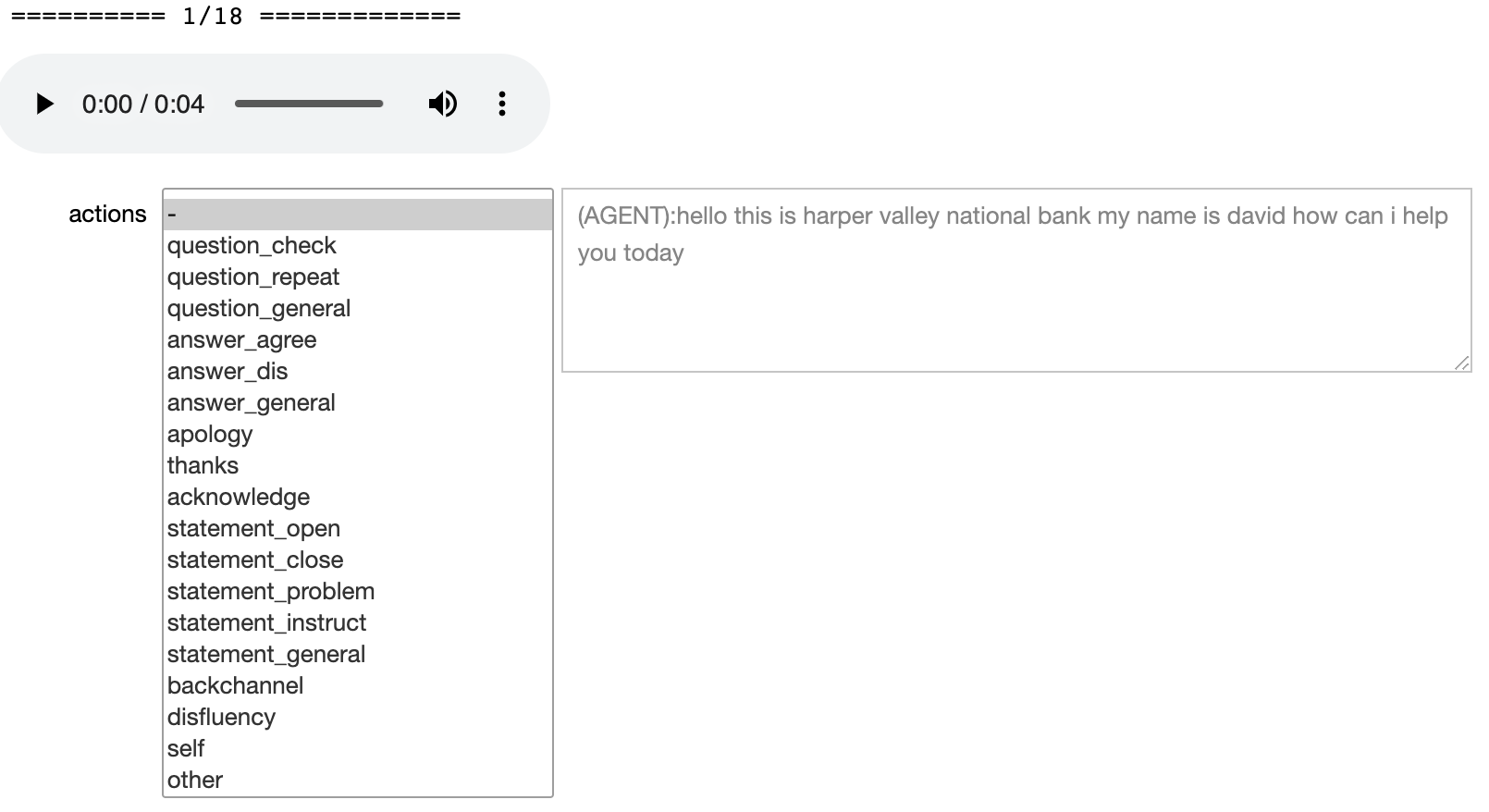}
    \caption{DAC annotation tool interface}
    \label{fig:dac_anno_tool}
\end{figure}

Figure~\ref{fig:dac_anno_tool} shows the annotation interface for DAC. Annotator could choose multiple acts per utterance. The annotator could listen to the corresponding speech segment for better judgment. Utterances are provided in chronologically by combining agent and caller channels. A single conversation was annotated by a single annotator. The total conversation was divided into 40 shards with evenly distributed intent of the conversation. A total of 5 annotators completed the annotation and we did not collect personal information such as the demographic or geographic background of the annotator.\footnote{annotators generally follows the principles here: https://datapractices.org/manifesto/\#principles}

\subsection{Model training details}
The E2E model fine-tuning was done with 2e-05 learning rate, 50,000 maximum update step and 2,800,000 maximum tokens for mini-batch. We use the macro-f1 score of dev set to choose the final model evaluation. We use single RTX GPU and took 2hours. Model training was done with 5 different random seed and reported median model. For pipeline system, wav2vec2 ASR model fine-tuning took 10 hours and DeBERTa NLP model took 3 hours using the same GPU. We followed the ASR and NLP fine-tuning script in SLUE-Toolkit. Reproducible baseline scripts will be released.

\subsection{Additional results}
\begin{figure}[htp!]
    \centering
    \includegraphics[width=0.6\linewidth]{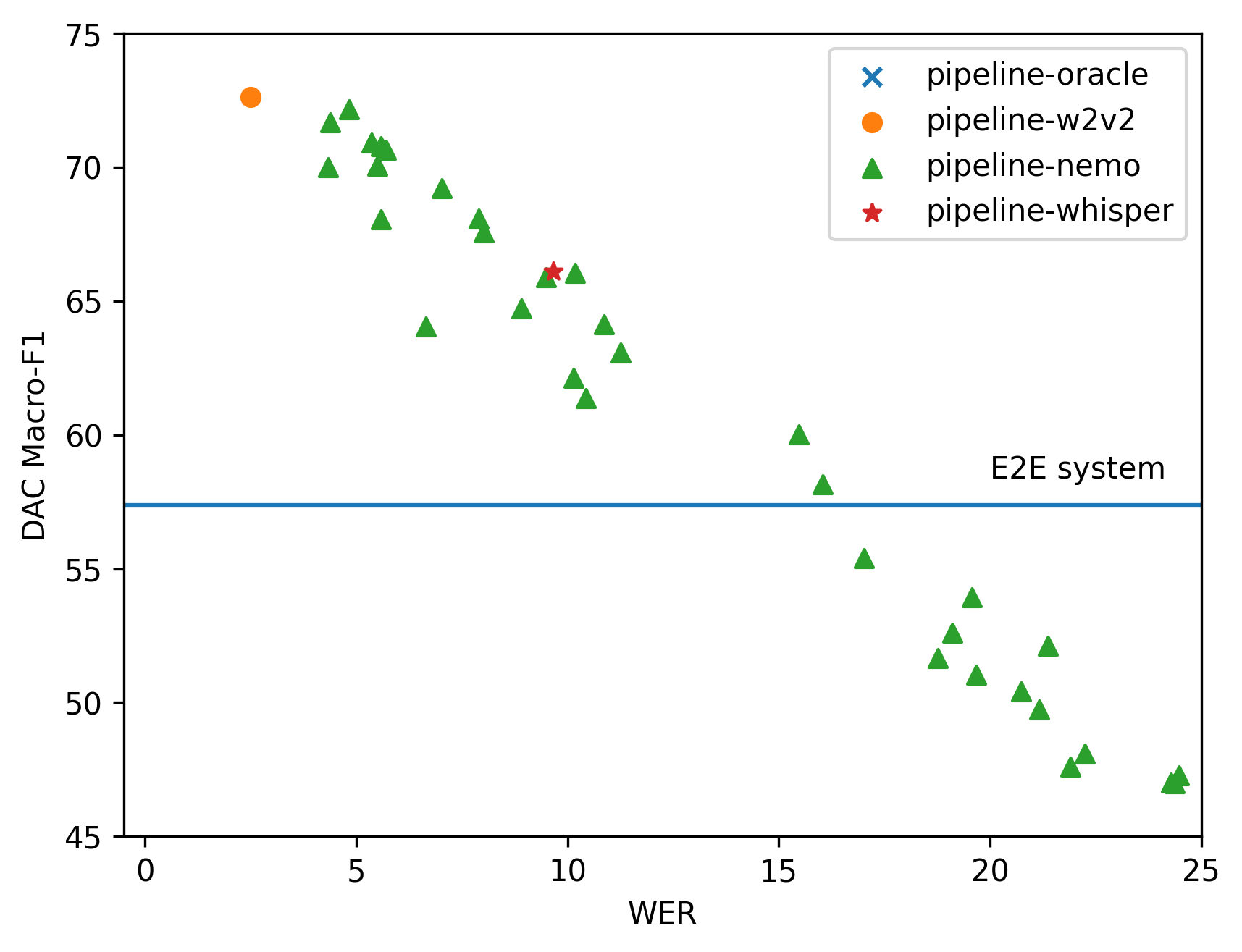}
    \caption{DAC task: WER and F1 scores on dev set}
    \label{fig:dac_corr_dev}
\end{figure}

Figure~\ref{fig:dac_corr_dev} shows the WER and F1 score on dev set and it shows the same trend compared to test set presented in Figure~\ref{fig:dac_corr}. Table~\ref{tab:dac_baseline_full} shows DAC task performance evaluation including dev and test set.

\begin{table}[htp!]\centering
\caption{DAC task baseline performance. *the best NeMo model based on DAC F1 score is "conformer-transducer-xxlarge"}\label{tab:dac_baseline_full}
\resizebox{10cm}{!}{%
\begin{tabular}{lcccc}\toprule
\multirow{2}{*}{System} &\multirow{2}{*}{\makecell{Speech\\model}} &\multirow{2}{*}{\makecell{Text\\model}} &\multicolumn{2}{c}{F1 score (WER)} \\\cmidrule{4-5}
& & &Dev &Test \\\midrule
pipeline-oracle &x &DeBERTa &76.1 (0.0) &72.3 (0.0) \\ \midrule
pipeline-w2v2 &wav2vec2 &DeBERTa &72.6 (2.5) &70.7 (2.1) \\
pipeline-nemo & best model* &DeBERTa &72.2 (4.8) &69.1 (4.8) \\
pipeline-whisper &whisper-en &DeBERTa & 66.1 (9.7)& 65.8 (8.1)\\ \midrule
E2E-w2v2 &wav2vec2 &x &57.4 (----) &57.9 (----) \\
\bottomrule
\end{tabular}
}
\end{table}

\newpage
\section{QA}
\subsection{Spoken question collection details}
\label{ssec:question-collection}
 
\begin{figure}[htp!]
\centering
    \includegraphics[width=0.7\linewidth]{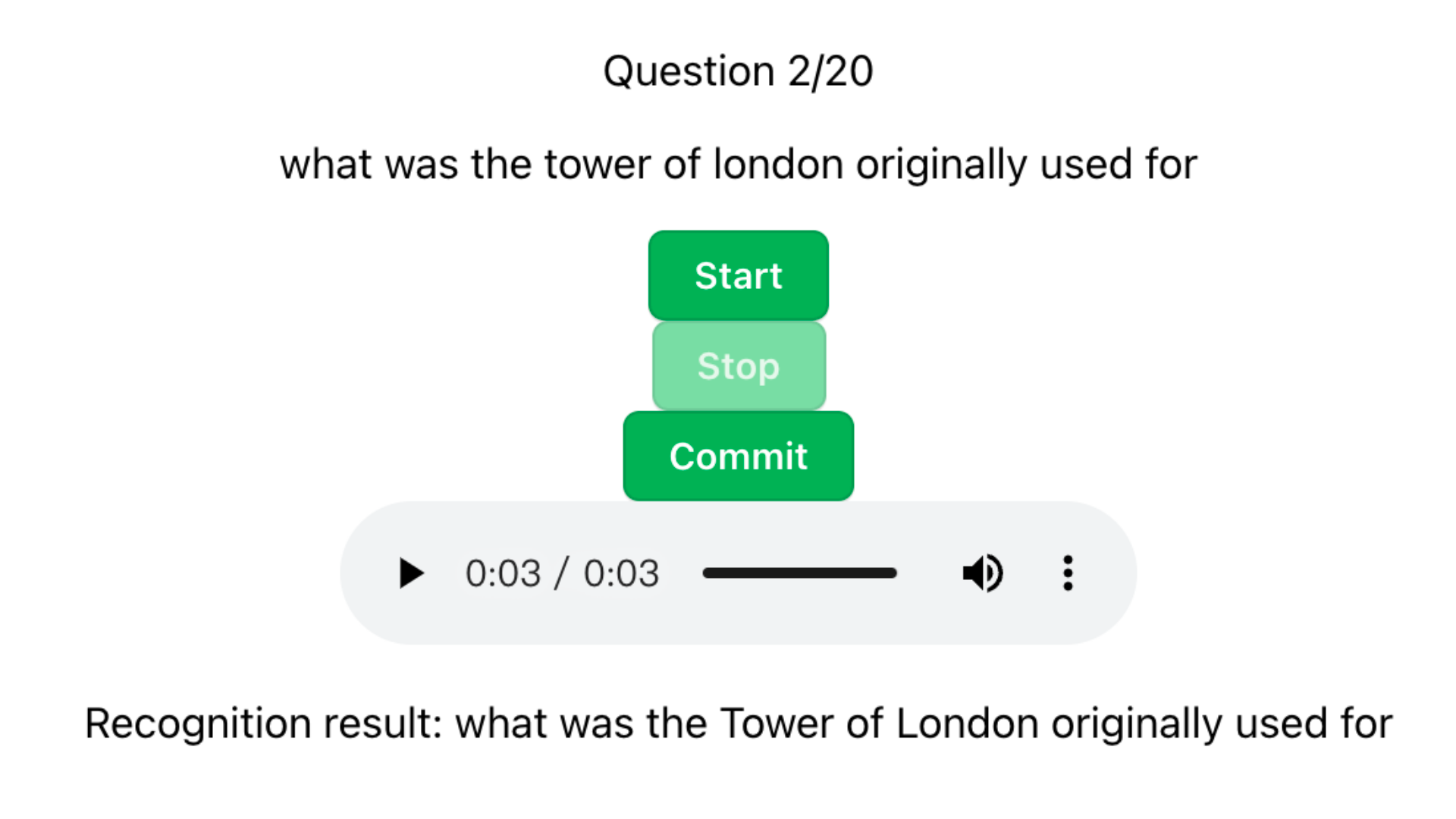}
    \caption{Interface of the website for spoken question collection in SLUE-SQA-5 dataset.}
    \label{fig:qa-website-interface}
\end{figure}

 To collection spoken questions in SLUE-SQA-5, we posted our own speech collection website to Mturk, asked each worker to read 50 questions and paid them 1 dollar, so the worker got 2 cents for reading one question.
 After the worker record their speech, our speech collection website uses Google Speech-to-Text service to transcribe the audio to text and calculate the WER. If the WER is higher than 30\%, our website will notify the worker and suggest them recording again. In our manual check, we listened to every recording by ourselves and discarded a recording only when we found that a high portion of the content was missing; otherwise, we still accepted it even if the WER was over 30\%. The interface of our speech collection website is shown in Figure~\ref{fig:qa-website-interface}.

\subsection{Search criteria of SLUE-SQA-5 documents}
\label{ssec:document-search}
When searching for the paired document to each question, we determined whether a document is relevant to a question by jointly considering (1) its rank among all documents in BM25~\citep{robertson2009probabilistic} search, a common term-based retrieval algorithm that scores the relevance between texts via keyword matching, (2) its rank among all documents in semantic search with the sentence-transformers model\footnote{\scriptsize {https://huggingface.co/sentence-transformers/multi-qa-mpnet-base-dot-v1}}~\citep{reimers2019sentence}, a neural sentence-level semantic encoder pre-trained on 215M QA pairs from multiple datasets, and (3) word-F1 derived by passing the question and the document through three different text QA models\footnote{\scriptsize {https://huggingface.co/Palak/microsoft\_deberta-large\_squad}}\footnote{\scriptsize {https://huggingface.co/deepset/deberta-v3-large-squad2}}\footnote{\scriptsize {https://huggingface.co/deepset/deberta-v3-base-squad2}} fine-tuned on SQuAD dataset.
We discard a question if we found no relevant document for it.

In specific, for each question, we searched for documents that meet all the criteria listed below:
\begin{itemize}
\item The document transcript includes the answer string to the question.
\item The document has one of the top-1000 highest BM25 scores with the question among all documents.
\item The document has one of the top-100 highest relevance scores with the question among all documents in semantic search with the sentence-transformers model.
\item When we pass the question and document through the three pre-trained text QA models mentioned in Section~\ref{sssection:qa-dataset}, at least one model gets a non-zero word-F1 score. (This criterion is used for dev and test set questions only.) 
\end{itemize}
If there exists a document that meet all the above criteria, we combine the document, question, and the question's answer into a question-answer-document triplet. 
%Then we get the start and end time of the answer span with Montreal Forced Aligner.
Otherwise, we consider the question unanswerable and discard it.
Note that we limit the number of paired document per question to one. If we find multiple documents that meet the criteria, we will choose the one with highest relevance score in semantic search among them as the paired document.

\subsection{Model training details}
\label{ssec:qa-model-training details}
The E2E-DUAL model is composed of a wav2vec2-large model encoding speech waveforms, a k-means model converting wav2vec2 layer representations into cluster IDs, and a Longformer model taking cluster IDs as input and predicting the start and end index of answer spans.
We extract the representations of Librispeech~\citep{panayotov2015librispeech} train-clean-100 set from the 22nd layer of the fixed wav2vec2-large model to train the k-means model.
The k-means model is then used to convert the representations of SLUE-SQA-5 fine-tune set into discrete units, which are taken as the input to the Longformer model.
 We fine-tune Longformer with 1e-4 learning rate, 500 warmup steps and overall 128 batch size on 4 Tesla V100 gpus. It takes around 40 hours to fine-tune the Longformer model for 45 epochs. 
 The total number of tuned parameters in DUAL, including the k-means model and Longformer part, is reported in Table~\ref{tab:models_size}.
 
For the pipeline system, we fine-tune the wav2vec2 ASR model with 1e-4 learning rate and 16 batch size for 10 epochs, and fine-tune the DeBERTa NLP model with 4e-5 learning rate, 100 warmup steps and 64 batch size for 10 epochs.
Wav2vec2 ASR model fine-tuning takes 25 hours and DeBERTa NLP model takes 6.5 hours using one V100 gpu.

\subsection{Additional results}
\label{ssec:qa-additional-results}
\begin{figure}[htp!]
\centering
    \includegraphics[width=0.7\linewidth]{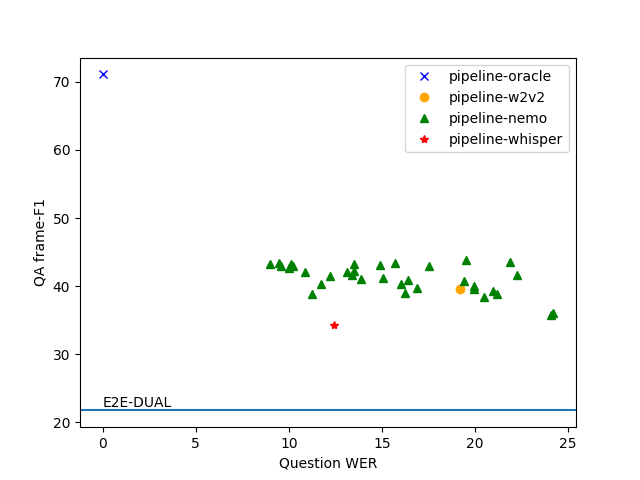}
    \caption{QA task: Question WER and frame-F1 scores}
    \label{fig:qa_corr_q}
\end{figure}

Figure~\ref{fig:qa_corr_q} shows the relationship between the question WER and frame-F1 on the test set.
We observe relatively weak correlation between question WER and frame-F1 compared to that between document WER and frame-F1.

\begin{figure}[htp!]
\centering
    \includegraphics[width=0.7\linewidth]{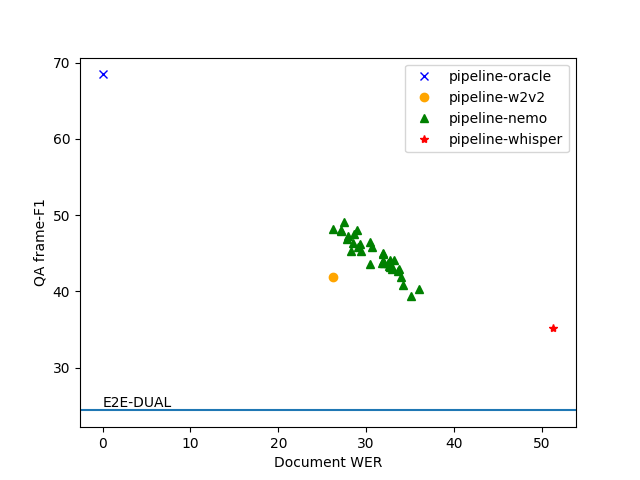}
    \caption{QA task: Document WER and frame-F1 scores on the dev set}
    \label{fig:qa_corr_d_dev}
\end{figure}

\begin{figure}[htp!]
\centering
    \includegraphics[width=0.7\linewidth]{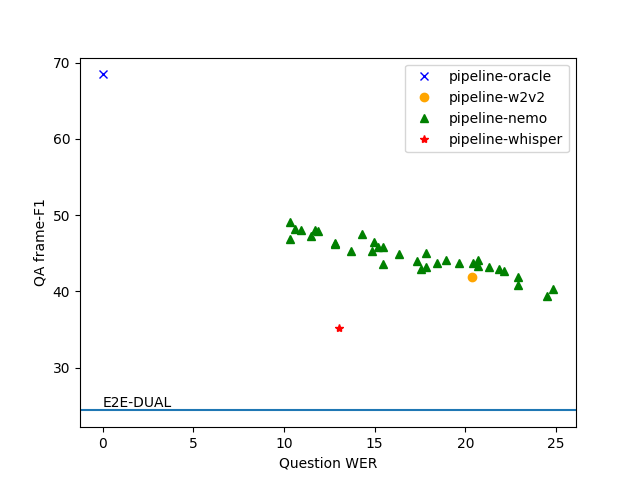}
    \caption{QA task: Question WER and frame-F1 scores on the dev set}
    \label{fig:qa_corr_q_dev}
\end{figure}

Table~\ref{tab:qa_baseline_dev} shows the QA performance on the dev set.
Figure~\ref{fig:qa_corr_d_dev} shows the relationship between document WER and frame-F1 on the dev set and has the similar trend (Pearson correlation coefficient=-0.94, p-value<0.01) compared to the test set in Figure~\ref{fig:qa_corr_d}.
Figure~\ref{fig:qa_corr_q_dev} shows the relationship between question WER and frame-F1 on the dev set.
Similar to the test set, we observe relatively weak correlation between question WER and frame-F1 compared to that between document WER and frame-F1.

\begin{table}[!htp]\centering
\caption{Number of SLUE-SQA-5 questions from each source text QA datasets.} \label{tab:num-questions}
\begin{tabular}{l|r|r|r|r|r|r}\toprule
& SQuAD & NQ & TriviaQA & WQ & TREC & total  \\\midrule
fine-tune & 11,900 & 12,383 & 20,358 & 1063 & 482 & 46,186 \\
dev & 679 & 85 & 869 & 212 & 94 & 1,939 \\
test &  828 & 125 & 1,051 & 266 & 112 & 2,382 \\
verified-test & 185 & 20 & 135 & 43 & 25 & 408 \\
\bottomrule
\end{tabular}
\end{table}

\begin{table}[!htp]\centering
\caption{QA task baseline performance on the dev set. *the best Nemo model based on frame-F1 score is "stt-en-contextnet-1024".}
\label{tab:qa_baseline_dev}
\resizebox{7.5cm}{!}{%
\begin{tabular}{lccc}\toprule
\multirow{2}{*}{System} &\multirow{2}{*}{\makecell{Speech\\model}} &\multirow{2}{*}{\makecell{Text\\model}} &\multicolumn{1}{c}{Frame-F1 } \\\cmidrule{4-4}
& & & Dev \\\midrule
pipeline-oracle &x & DeBERTa & 68.5  \\\midrule
pipeline-w2v2 & wav2vec2 & DeBERTa & 41.8   \\
pipeline-nemo & best model* & DeBERTa &  49.2  \\
pipeline-whisper &whisper-en & DeBERTa & 35.2  \\\midrule
E2E-DUAL & DUAL & x & 24.4  \\ 
\bottomrule
\end{tabular}
}
\end{table}

\newpage
\section{SUMM}
\subsection{Model details}
\label{sec:summ-appendix-hyperparam}
The ASR models consist of a conformer encoder-decoder architecture with pre-trained SSL representations like Hubert large~\cite{hsu2021hubert} and wav2vec2 large~\cite{baevski2020wav2vec} representations as features. Following prior work~\cite{pretrain-E2E-SLU}, a weighted sum of multiple hidden states of SSL models is utilized. Since the TED talks are very long, we break the audio into 10 second chunks, and infer the most likely transcript for each chunk independently. Then we concatenate the resulting transcripts from each audio chunk to obtain the talk transcript. ASR models were trained for nearly 23 hours on 4 v100 gpus.

The E2E speech summarization model has similar architecture as the ASR model of the pipeline baseline. Since the TED talks were too long to fit the entire speech input on a GPU, we use only the last hidden state of SSL model and trained our E2E model using only the first 30000 speech frames (600 seconds). E2E speech summarization model was trained for nearly 16 hours on 4 v100 gpus.

For Nemo conformer and squeezeformer models, the audio is too long to perform inference using a GPU, and hence we have to break audio input into 5-minute chunks and perform inference separately on each of these chunks.
\subsection{Additional dataset details}
\label{sec:summ-data-details}
\begin{figure}[hbp!]
\centering
\begin{subfigure}[b]{\linewidth}
\centering
\includegraphics[width=0.9\linewidth]{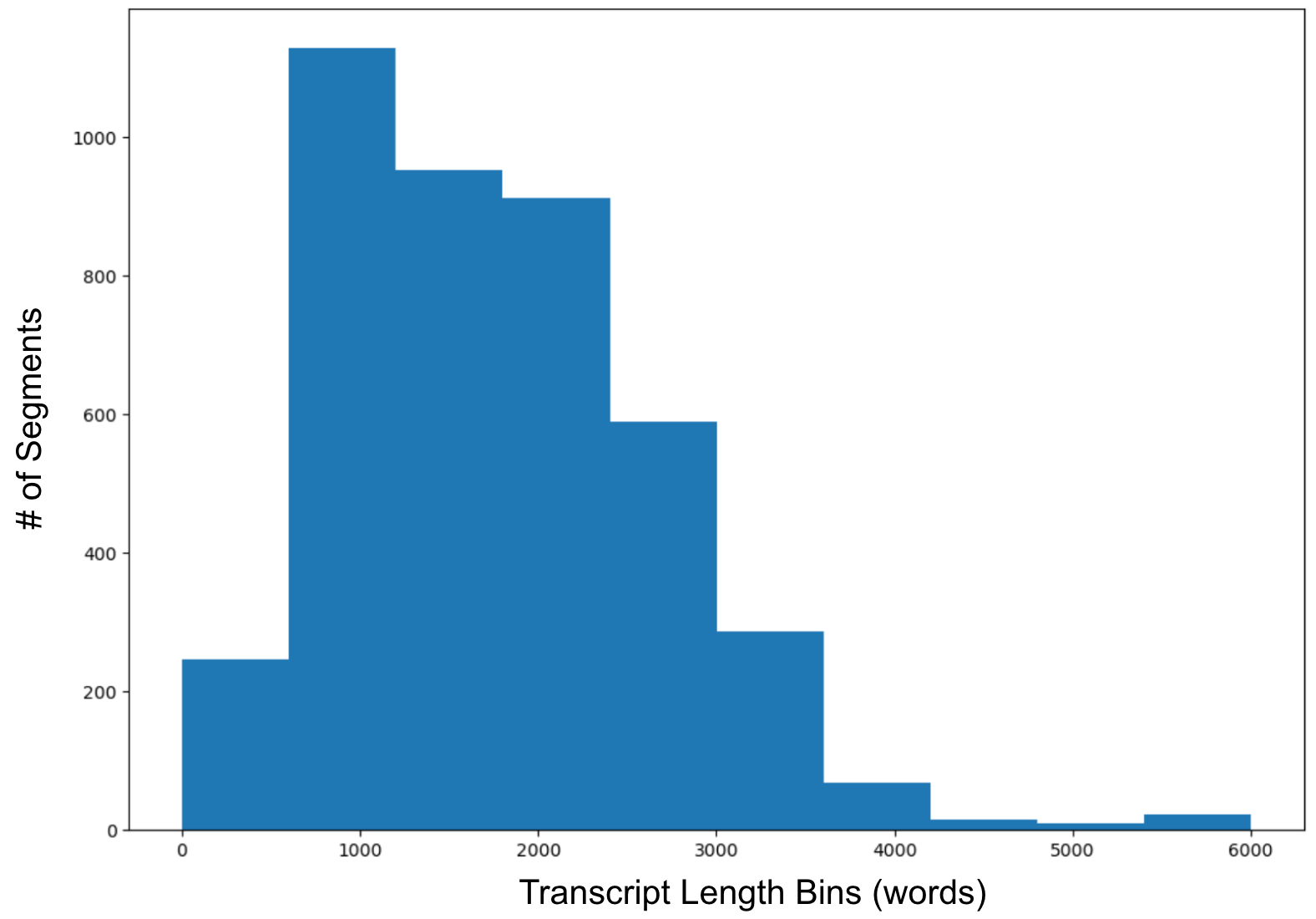}
\caption{Transcript length distribution}
\label{fig:transcript_length_dist}
\end{subfigure}
\begin{subfigure}[b]{\linewidth}
\centering
\includegraphics[width=0.9\linewidth]{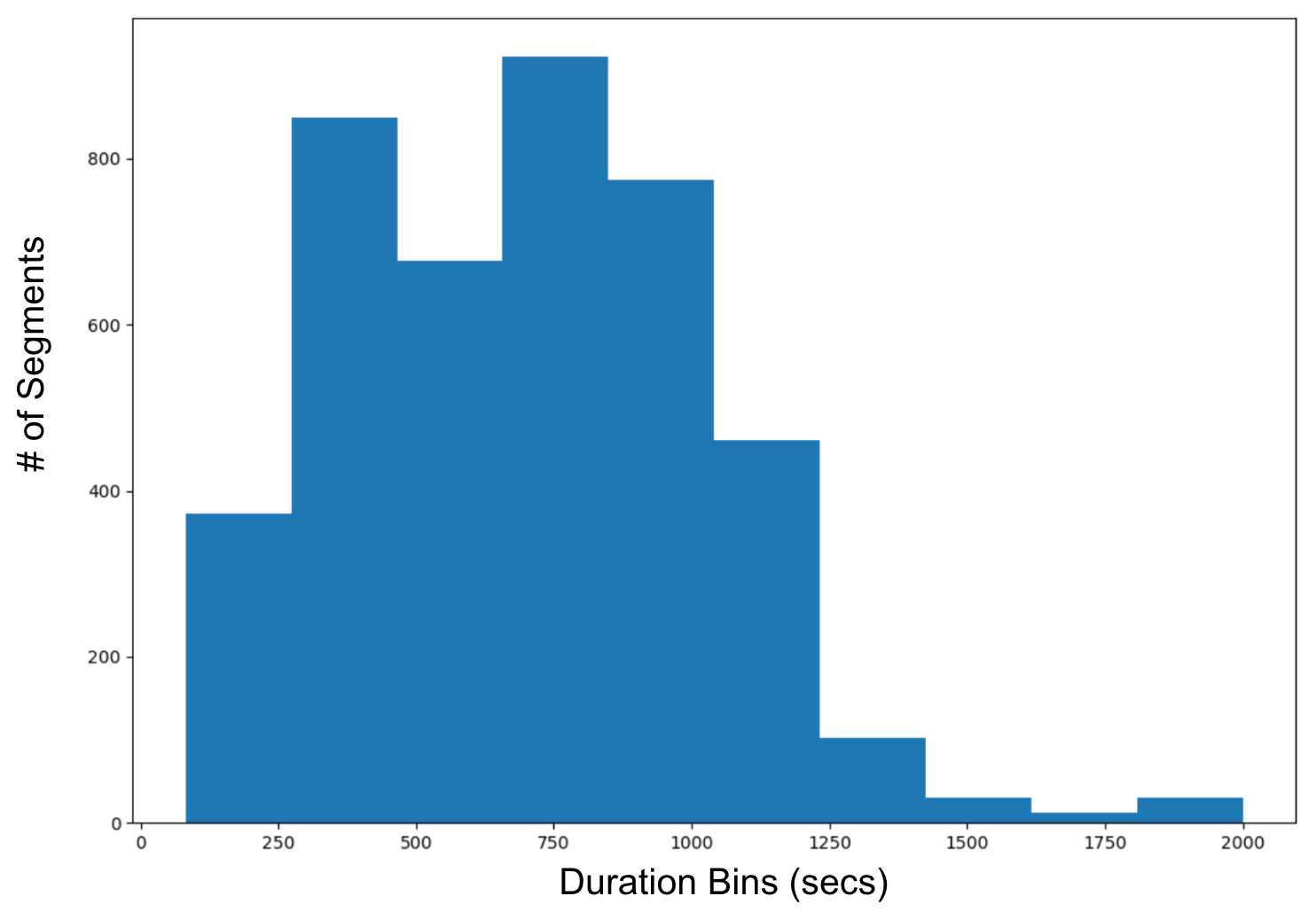}
\caption{Audio duration distribution}
\label{fig:duration_dist}
\end{subfigure}
\caption{Figure showing transcript length and audio duration distribution in TED Summary dataset}
\label{fig:summary-dist}
\end{figure}

Table~\ref{tab:ted_stats} summarizes the statistics of the dataset, and the distribution of ground truth transcript and summaries is shown in Figure ~\ref{fig:summary-dist}. We observe that this dataset contains much longer audios and transcripts than prior works.
\begin{table*}[!htp]\centering
\caption{SLUE-TED data statistics}\label{tab:ted_stats}
\resizebox{\linewidth}{!}{
\begin{tabular}{l|r|r|r|r|r}\toprule
Corpus & utterances & duration (h) & duration/utt (s) & Transcript length (words) & Summary length (words)\\\midrule
How2 & 79114 &  1890 & 86 & 853 & 60\\
SLUE-TED & 4233 & 829 & 705 & 1757 & 61\\
\bottomrule
\end{tabular}}
\end{table*}

\subsection{Additional results}
\label{sec:summ-res-details}
\begin{figure}[ht]
\centering
    \includegraphics[width=0.5\linewidth]{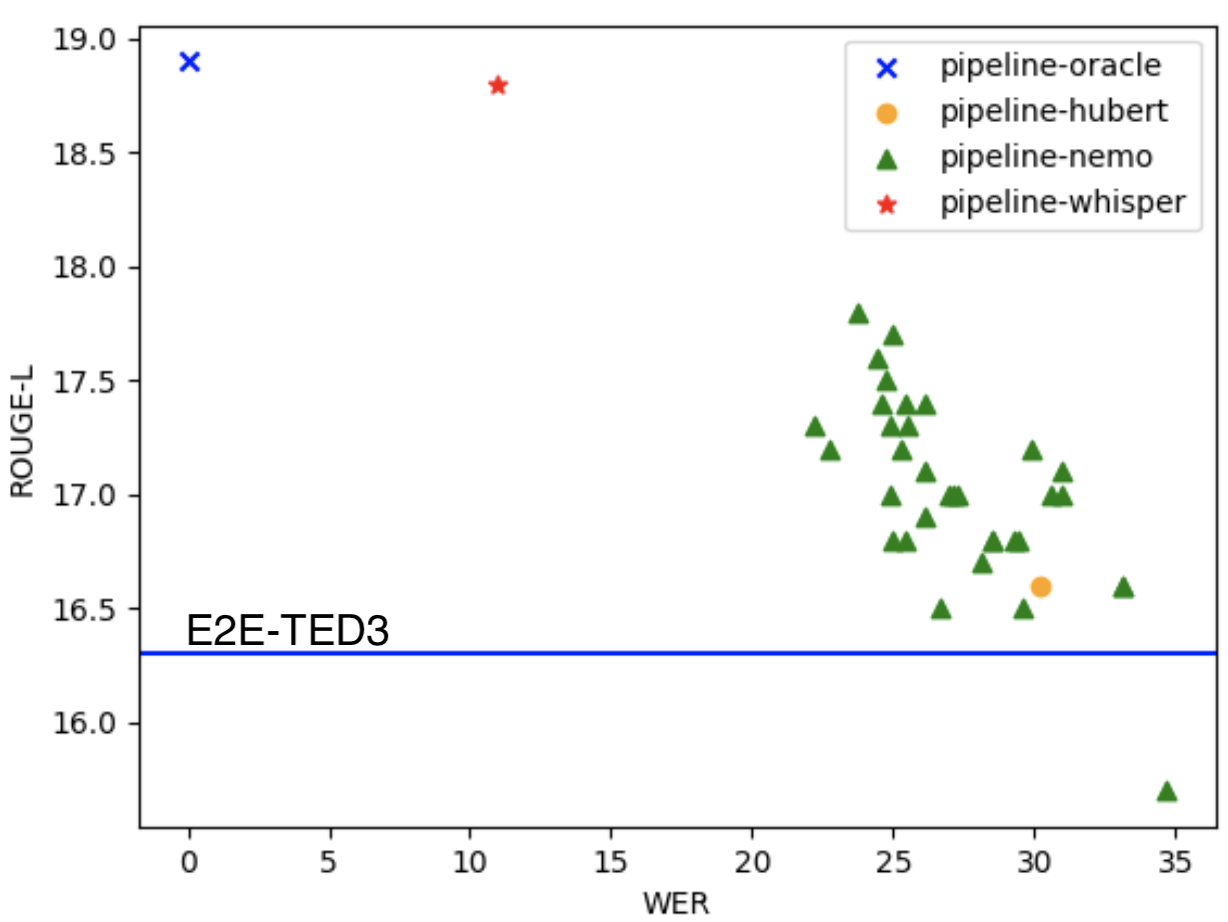}
    \caption{SUMM task : WER and ROUGE-L score on dev set}
    \label{fig:summ_valid_corr}
\end{figure}

Table~\ref{tab:summ_valid_baseline} shows the performance of all the models on the dev set. Figure~\ref{fig:summ_valid_corr} shows the correlation between WER and ROUGE-L scores on the dev set and has a similar trend to the one observed on test set in figure~\ref{fig:summ_corr}. Table~\ref{tab:exact:matches} show the percentage of exact matches in reference summary and predicted summaries for each POS tag on the test set. We further analyzed the performance of our E2E Summ model separately on abstract and title in summary and observed that the model performs slightly better at generating title (ROUGE-L:15.2, BERTScore:87.7) as compared to generating the abstract (ROUGE-L:14.4, BERTScore:83.4). Table~\ref{tab:anecdotes}
provides example summaries generated by our baseline systems. We observe that pipeline models generate more accurate words while E2E model generates more semantically similar summaries to reference. However, both these models generate summaries that differ from references suggesting significant room for improvement. 
\begin{table*}[!htp]\centering
\caption{SUMM task baseline performance on the dev set. The ASR models are trained on the TEDLIUM-3 corpus. For pipeline models, we also experiment with training NLU model on ASR Transcripts (ASR) instead of ground truth transcript. *the best nemo model based on SUMM ROUGE-L score is "conformer-transducer-xxlarge".}\label{tab:summ_valid_baseline}
\resizebox{15cm}{!}{%
\begin{tabular}{lccccccccc}\toprule
System & \makecell{Speech\\model} & \makecell{Text\\model} & ROUGE-1 & ROUGE-2 & ROUGE-L & METEOR & BERTScore & WER \\
\midrule
pipeline-oracle &x & LED & 29.4 & 7.2 & 18.9 & 13.3 & 83.7 & 0.0\\\hline
pipeline-wv2v2 & W2V2-ASR & LED & 26.7 & 5.5 & 17.0 & 12.2 & 82.6 & 34.5\\
pipeline-hubert & Hubert-ASR & LED & 26.6 & 5.3 & 16.6 & 12.3 & 82.5 & 30.2\\
pipeline-nemo & best model* & LED & 27.4 & 5.8 & 17.3 &  12.7 & 82.6 & 25.5\\
pipeline-whisper & whisper-en & LED & 29.1 & 7.2 & 18.8 & 13.1 & 83.7 & 11.0\\
pipeline-whisper ASR & whisper-en & LED(ASR) &  29.1 & 7.3 & 18.9 &     13.3 & 83.7 & 11.0\\ \hline
E2E-TED3 & TEDLIUM3-Conformer & x & 23.9 & 5.2 & 16.3 &  10.4 & 84.3 & ---\\
\bottomrule
\end{tabular}
}
\end{table*}

\begin{table}[!htp]\centering
\caption{Matches in predicted summary and reference summary for different POS tags}\label{tab:exact:matches}
{%
\begin{tabular}{c|c}\toprule
POS Tag & Matches(\%) \\\midrule
PROPN & \hphantom{0}6.1\\
AUX & 42.5\\
ADJ & 10.8\\
CCONJ & 55.1\\
ADV & \hphantom{0}9.7\\
VERB & 11.3\\
PRON & 34.3\\
NOUN & 19.7\\
DET & 82.5\\
\bottomrule
\end{tabular}
}
\end{table}

\begin{table}[t]
\caption{SLUE-TED Summarization examples.  }.
\label{tab:anecdotes}
  \centering
\resizebox {\linewidth} {!} {
\begin{tabular}{lp{155mm}}
\toprule
Method     & Example    \\ \midrule
Reference &  The work that makes all other work possible [SEP] Domestic workers are entrusted with the most precious aspects of people's lives -- they're the nannies, the elder-care workers and the house cleaners who do the work that makes all other work possible. Too often, they're invisible, taken for granted or dismissed as "help"; yet they continue to do their wholehearted best for the families and homes in their charge. In this sensational talk, activist Ai-Jen Poo shares her efforts to secure equal rights and fair wages for domestic workers and explains how we can all be inspired by them. "Think like a domestic worker who shows up and cares no matter what" she says. \\
pipeline-hubert & The domestic workers' rights movement [SEP] In the US, domestic workers are often characterized as unskilled, unskilled and largely uneducated -- a legacy that's often cast aside for more humane work. But in this bold, human talk, Ameera Al-Sabouni advocates for a new kind of work, one that includes days of rest, paid time off and other protections for domestic workers -- and shares how the movement for domestic workers' rights is gaining legislative momentum.\\
E2E-TED3 & The work that makes all other work possible? [SEP] What makes all other work possible? In this world, it's possible, says important immorality domestic workers are so fundamental to the very basics of our lives, says lawyer and lawyer and TED Fellow Juan Enriquez. She tells the story of how workplaces that makes all other work possible.\\ \midrule
Reference & The link between fishing cats and mangrove forest conservation [SEP] Mangrove forests are crucial to the health of the planet, gobbling up CO2 from the atmosphere and providing a home for a diverse array of species. But these rich habitats are under continual threat from deforestation and industry. In an empowering talk, conservationist and TED Fellow Ashwin Naidu shares how community-driven efforts in South and Southeast Asia are working to protect mangroves -- all with a little help from the mysterious and endangered fishing cat. \\
pipeline-hubert & Why protecting forests is the best thing for the environment [SEP] protecting one acre of rainforests in south east asia may well be like protecting five or more acres of tropical forests in the future. But would you like to eliminate your entire life's carbon footprint? Eco-entrepreneur and TED fellow Sophia Kianni considers that action is being taken to protect these precious ecosystems -- and the millions of people who live next to them. "Mangroves are more than just their home to a fast-growing ecosystem" she says. "They can be the first line of defense between storm surges, tsunamis and the millions of people who live next to these forests for their survival."\\
E2E-TED3 & The tigers of the Mangroves [SEP] We can all be part of a future where fishing cats are threatened by habitat loss, loves to fish and lives in some of the most unique and valuable ecosystems on earth, mainly because of our international deforestations, local people and the global community. So what's learned that we can all be part of a future where fishing cats are threatened by habitat loss, local people and the global community. In this eye-opening talk, she shares how these restored Mangroves may be lost.\\ 
\bottomrule
\end{tabular}
}
\end{table}
\clearpage
\section{Named entity localization}
\label{sec:nel-appendix}
\subsection{Annotation details}
\label{sec:nel-appendix-annot}
As described in Sec.~\ref{sec:nel-data-details}, we use MFA to obtain ground-truth word-level alignments. When we run MFA, it fails to align twenty-six files across dev and test splits. On manual inspection we identify differences in audio utterance and the corresponding text transcript due to incorrect end-pointing for twenty-two of these files. These cases have contiguous words at the end of the transcript that are not a part of the audio utterance. Running MFA after removing these extra words from the transcripts fixes these cases. But, for seven of these files, at least one entity word is a part of the missing words and so, the time alignments don't have all the entity phrases that are a part of the published SLUE-NER annotations. In the interest of utterance-level consistency between SLUE-NER and SLUE-NEL, we skip these files. For the remainder four of the twenty-six files that MFA fails to align, we manually add the word alignments using Praat software~\cite{Boersma2009}.

In order to check the validity of MFA produced alignments, we manually verify the entity alignments for 372 entity phrases across randomly chosen 188 utterances in dev split. This constitutes 20\% of all entity phrases in the dev split and thus our analysis should be representative for the complete split. Our manual pass exposed 51 of 372 phrases to be misaligned and the nature of misalignment varied from a minor offset to being completely off. In order to quantify the effect of the identified misalignments on our evaluation metrics, we manually rectify the alignments for these 51 phrases and report the following scores for this representative set of 188 utterances:
\begin{enumerate}[leftmargin=*,noitemsep,nolistsep]
    \item The frame-F1 between rectified and original timestamps is 96\%,
    \item The relative difference in baseline model scores (evaluating models listed in Table~\ref{tab:nel_baseline}) using these two versions as ground-truths is <3\%,
    \item The general trend in baseline model scores is similar across models for the results using these two versions as ground-truths.
\end{enumerate}

Thus, we conclude that the alignments produced by MFA are reliable for robustly comparing between different modeling approaches and can be used as ground-truth despite minor issues in the generated time-stamps. Additionally, we find that the faulty timestamps are a result of imperfect transcripts in VoxPopuli and not an issue with MFA. The imperfections in these transcripts are expected, since the data is originally curated with 20\% character error rate threshold~\cite{wang2021voxpopuli}.

\subsection{Hyperparameter details}
\label{sec:nel-appendix-hyperparam}
NEL evaluation has two hyperparameters,{\emph offset} and \emph{incl\_blank}. We evaluate the dev set on a range of offset values between -0.3 seconds and 0.3 seconds with an increment of 20 milliseconds. The \emph{incl\_blank} is a Boolean hyperparameter. The best hyperparameter values based on dev set performance are listed in Table~\ref{tab:nel_bestparams}.

The 34 NeMo models have one of the three types of decoding strategies -- (i) character-level CTC, (ii) subword-level CTC, and (iii) subword-level RNN transducer (RNNT).
The character-level CTC models are processed in the same way as the {\it pipeline-w2v2} models, where the \emph{incl\_blank} denotes whether or not the $\epsilon$ tokens before and after the entity phrase, between the word separator tokens, are included in the entity time stamp. 
The subword-level CTC model vocabulary in the NeMo toolkit does not have a word separator token, and instead, the start of the word is characterized by an ``\_'' prepended to a subword. So, the \emph{incl\_blank} denotes whether the trailing $\epsilon$ tokens, before the start of the next word, are included in the entity time stamp.
The RNNT model class in the NeMo toolkit directly gives subword-level start times, so \emph{offset} was the only relevant hyperparameter here. 

\begin{table}[hbp!]
\caption{Best hyperparameters for NEL models}\label{tab:nel_bestparams}\centering
\resizebox{16cm}{!}{%
\begin{tabular}{lllll}\toprule
System & Speech model & \begin{tabular}[c]{@{}c@{}}Training\\ objective\end{tabular} & offset (s) & incl\_blank \\\midrule
E2E-w2v2 & wav2vec2 & char-CTC & 0.00 & True \\\hline
pipeline-w2v2 & wav2vec2 & char-CTC & -0.08 & True \\\hline
\multirow{3}{*}{pipeline-nemo} & QuartzNet15x5Base-En & \multirow{3}{*}{char-CTC} & -0.22 & True \\
  & stt\_en\_jasper10x5dr &   & -0.26 & True \\
  & stt\_en\_quartznet15x5 &   & -0.26 & True \\\hline
\multirow{6}{*}{pipeline-nemo} & stt\_en\_citrinet\_1024 & \multirow{6}{*}{subword-CTC} & -0.10 & True \\
  & stt\_en\_citrinet\_1024\_gamma\_0\_25 &   & -0.10 & True \\
  & stt\_en\_citrinet\_256 &   & -0.10 & True \\
  & stt\_en\_citrinet\_256\_gamma\_0\_25 &   & 0.00 & True \\
  & stt\_en\_citrinet\_512 &   & -0.12 & True \\
  & stt\_en\_citrinet\_512\_gamma\_0\_25 &   & -0.16 & True \\\hline
\multirow{7}{*}{pipeline-nemo} & stt\_en\_conformer\_ctc\_large & \multirow{7}{*}{subword-CTC} & -0.12 & True \\
  & stt\_en\_conformer\_ctc\_large\_ls &   & -0.02 & False \\
  & stt\_en\_conformer\_ctc\_medium &   & -0.12 & True \\
  & stt\_en\_conformer\_ctc\_medium\_ls &   & -0.02 & False \\
  & stt\_en\_conformer\_ctc\_small &   & -0.08 & True \\
  & stt\_en\_conformer\_ctc\_small\_ls &   & 0.00 & False \\
  & stt\_en\_conformer\_ctc\_xlarge &   & -0.08 & True \\\hline
\multirow{6}{*}{pipeline-nemo} & stt\_en\_squeezeformer\_ctc\_large\_ls & \multirow{6}{*}{subword-CTC} & -0.02 & False \\
  & stt\_en\_squeezeformer\_ctc\_medium\_large\_ls &   & -0.02 & False \\
  & stt\_en\_squeezeformer\_ctc\_medium\_ls &   & -0.02 & False \\
  & stt\_en\_squeezeformer\_ctc\_small\_ls &   & -0.02 & False \\
  & stt\_en\_squeezeformer\_ctc\_small\_medium\_ls &   & -0.02 & False \\
  & stt\_en\_squeezeformer\_ctc\_xsmall\_ls &   & -0.02 & False \\\hline
\multirow{6}{*}{pipeline-nemo} & stt\_en\_conformer\_transducer\_large & \multirow{6}{*}{subword-RNNT} & 0.16 & n/a \\
  & stt\_en\_conformer\_transducer\_large\_ls &   & 0.14 & n/a \\
  & stt\_en\_conformer\_transducer\_medium &   & 0.20 & n/a \\
  & stt\_en\_conformer\_transducer\_small &   & 0.20 & n/a \\
  & stt\_en\_conformer\_transducer\_xlarge &   & 0.18 & n/a \\
  & stt\_en\_conformer\_transducer\_xxlarge &   & 0.18 & n/a \\\hline
\multirow{6}{*}{pipeline-nemo} & stt\_en\_contextnet\_1024 & \multirow{6}{*}{subword-RNNT} & 0.22 & n/a \\
  & stt\_en\_contextnet\_1024\_mls &   & 0.30 & n/a \\
  & stt\_en\_contextnet\_256 &   & 0.14 & n/a \\
  & stt\_en\_contextnet\_256\_mls &   & 0.20 & n/a \\
  & stt\_en\_contextnet\_512 &   & 0.22 & n/a \\
  & stt\_en\_contextnet\_512\_mls &   & 0.30 & n/a \\
\bottomrule
\end{tabular}
}
\end{table}

\subsection{Error analysis}
\label{sec:nel-appendix-error}
Table~\ref{tab:nel_prec_recall_analysis} shows precision and recall values for the NEL models. The E2E model outperforms in {\it precision} (i.e, more predicted regions are named entities), whereas the pipeline model outperforms in {\it recall}. 
The mismatch in text NER's training (ground-truth text) and inference (ASR output) could lead to higher false positives in the pipeline model. 

\begin{figure}[hbp!]
    \centering
    \includegraphics[width=0.7\linewidth]{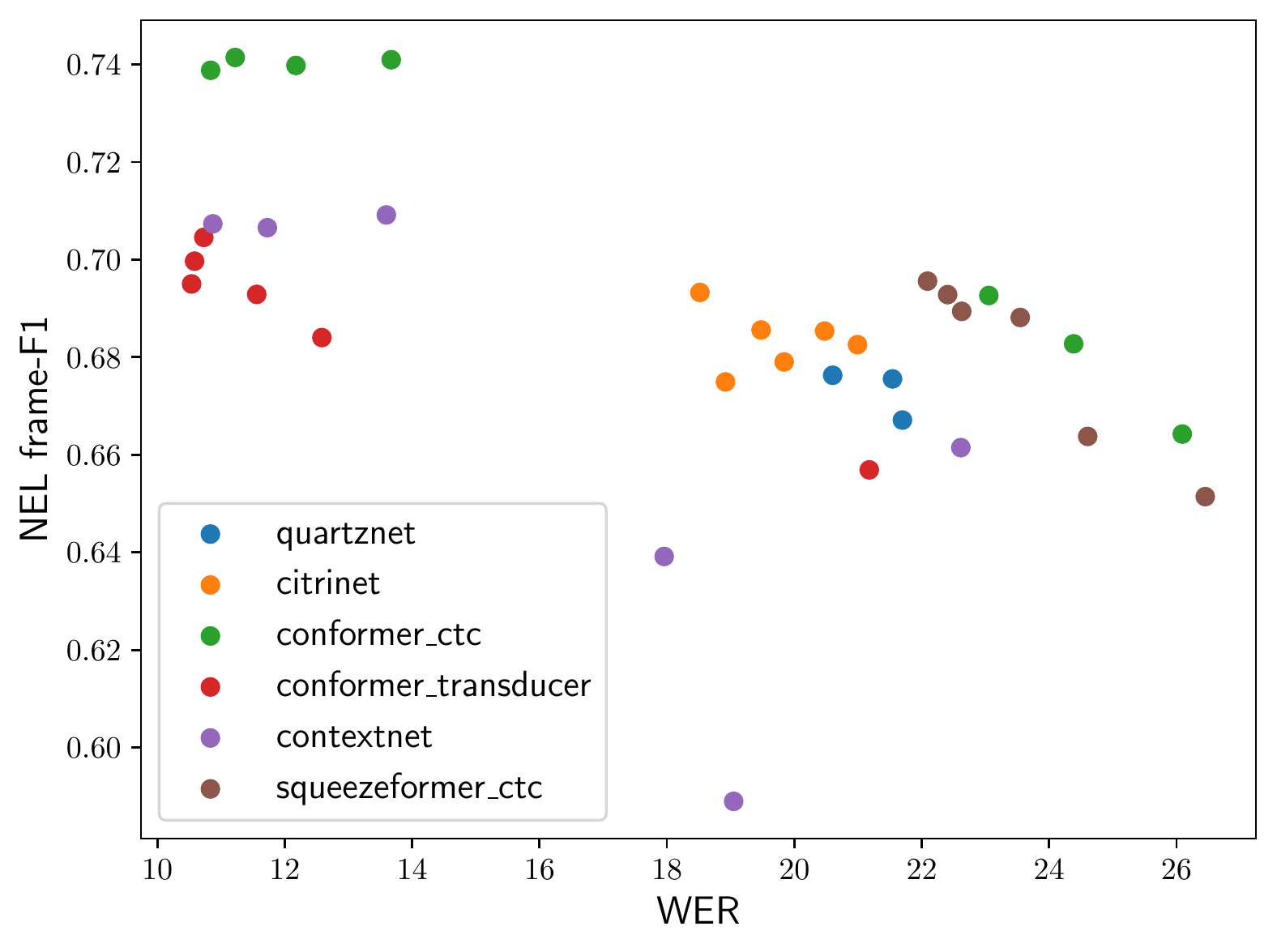}
    \caption{WER and frame-F1 scores on test set for different NeMo models}
    \label{fig:nel_nemo_corr}
\end{figure}

Figure~\ref{fig:nel_nemo_corr} shows the scatter plot between WER and F1 scores for NeMo, where the points are color-coded for different base model types. We see that the NEL and ASR performance are correlated within a single model category.

\begin{table*}[hbp!]\centering
\caption{NEL task baseline precision and recall performance on dev set. *the best nemo model based on NEL frame-f1 score on dev is ``stt\_en\_conformer\_ctc\_small"}
\label{tab:nel_prec_recall_analysis}
\resizebox{15cm}{!}{%
\begin{tabular}{lcc|cc|cc|cc|cc}\toprule
\multirow{2}{*}{System} &\multirow{2}{*}{\makecell{Speech\\model}} & \multirow{2}{*}{\makecell{Text\\model}} & \multicolumn{2}{c|}{frame-F1} & \multicolumn{2}{c|}{word-F1 ($\rho$=1)} & \multicolumn{2}{c|}{word-F1 ($\rho$=0.8)} & \multicolumn{2}{c}{word-F1 ($\rho$=0.5)} \\\cmidrule{4-11}
& & & Prec. & Recall & Prec. & Recall & Prec. & Recall & Prec. & Recall \\\midrule
pipeline-oracle & x & DeBERTa & 91.7 & 92.8 & 92.4 & 94.7 & 92.4 & 94.7 & 92.4 & 94.7 \\
pipeline-w2v2 & wav2vec2 & DeBERTa & 57.8 & 78.8 & 70.4 & 46.4 & 71.1 & 74.1 & 68.5 & 84.9 \\
E2E-w2v2 & wav2vec2 & x & 81.0 & 51.7 & 71.8 & 19.5 & 83.8 & 55.0 & 83.2 & 63.2 \\
pipeline-nemo & best model* & DeBERTa & 69.2 & 83.2 & 82.4 & 56.4 & 83.7 & 83.1 & 79.7 & 88.1 \\
\bottomrule
\end{tabular}
}
\end{table*}

\begin{figure}[hbp!]
    \centering
    \includegraphics[width=0.7\linewidth]{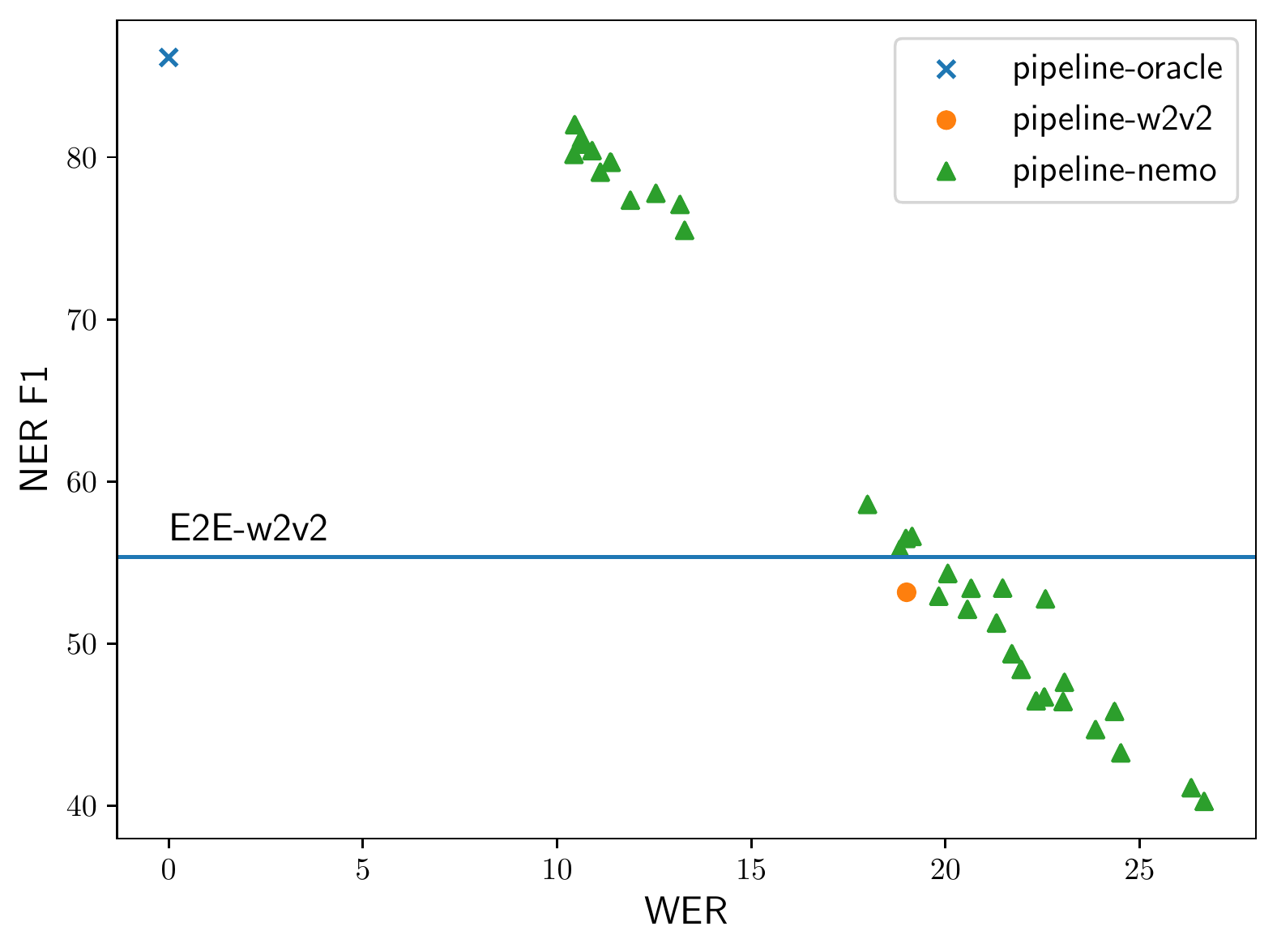}
    \caption{NER task: WER and frame-F1 scores on dev set}
    \label{fig:ner_corr}
\end{figure}

\subsection{Additional results}
\label{sec:nel-appendix-additional}
Table~\ref{tab:nel_baseline_all} shows performance of NEL for dev and test sets across different thresholds for word-F1. For word-F1, relaxing the tolerance from $\rho=1$ to $\rho=0.8$ gives a major performance boost -- up to 30\% and 116\% relative for pipeline and E2E models respectively.

\begin{table*}[hbp!]\centering
\caption{NEL task baseline performance. The wav2vec2 models are fine-tuned on slue-voxpopuli data.*the best NeMo model based on NEL frame-f1 score on dev is ``stt\_en\_conformer\_ctc\_small"
}
\label{tab:nel_baseline_all}
\resizebox{15cm}{!}{%
\begin{tabular}{lcc|cc|cc|cc|cc}\toprule
\multirow{2}{*}{System} &\multirow{2}{*}{\makecell{Speech\\model}} & \multirow{2}{*}{\makecell{Text\\model}} & \multicolumn{2}{c|}{frame-F1} & \multicolumn{2}{c|}{word-F1 ($\rho$=1)} & \multicolumn{2}{c|}{word-F1 ($\rho$=0.8)} & \multicolumn{2}{c}{word-F1 ($\rho$=0.5)} \\\cmidrule{4-11}
& & & Dev & Test & Dev & Test & Dev & Test & Dev & Test \\\midrule
pipeline-oracle & x & DeBERTa & 92.3 & 89.0 & 93.6 & 90.0 & 93.6 & 90.0 & 93.6 & 90.0 \\
pipeline-w2v2 & wav2vec2 & DeBERTa & 66.9 & 65.1 & 56.0 & 53.6 & 72.7 & 72.1 & 75.9 & 74.1 \\
E2E-w2v2 & wav2vec2 & x & 63.2 & 56.2 & 30.8 & 25.7 & 66.5 & 59.4 & 71.8 & 64.6 \\
pipeline-nemo & best model* & DeBERTa & 75.5 & 74.1 & 66.9 & 64.0 & 83.4 & 81.4 & 83.7 & 81.0 \\
\bottomrule
\end{tabular}
}
\end{table*}

\begin{figure}[h]
    \centering
    \includegraphics[width=0.8\linewidth]{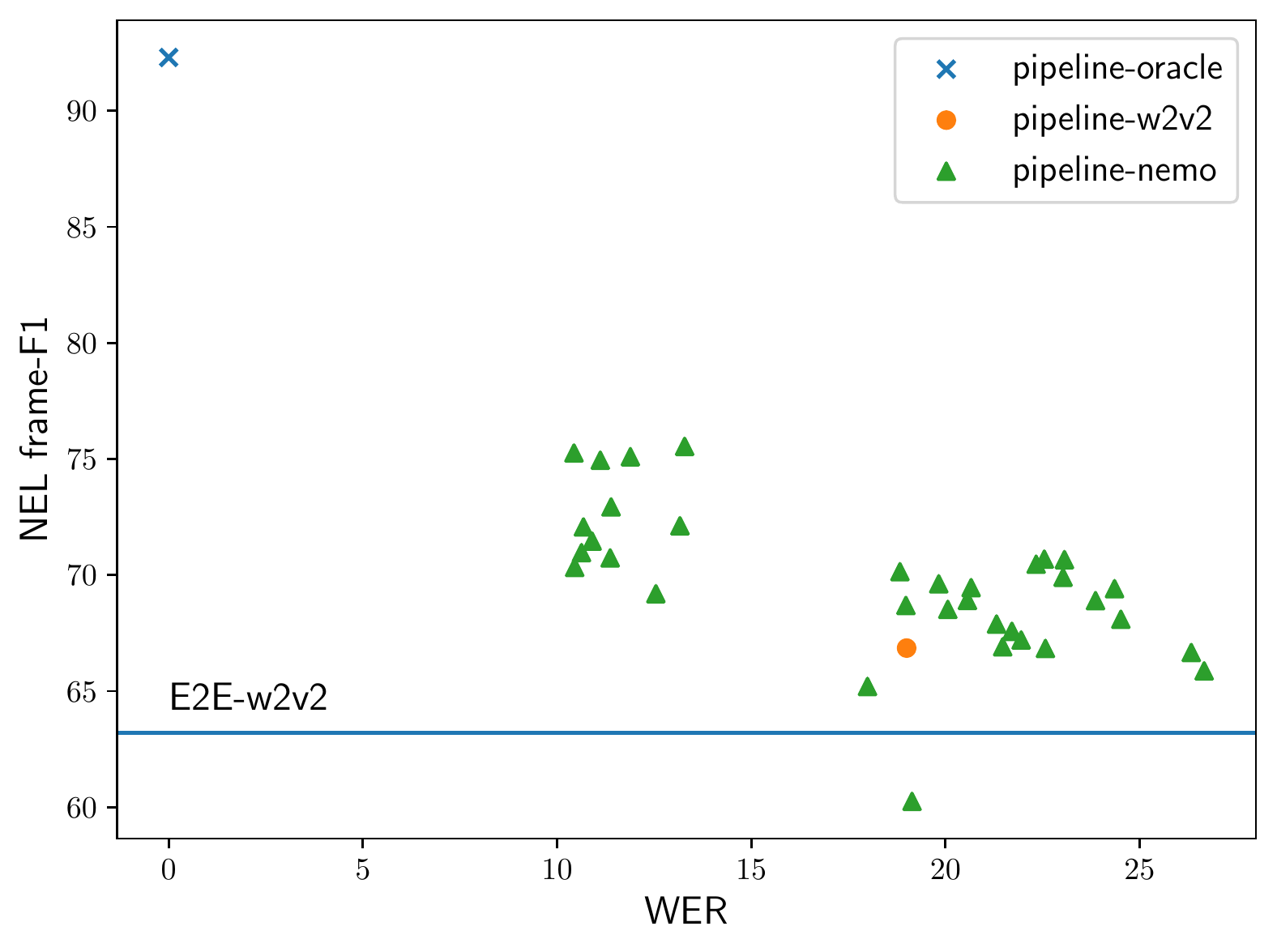}
    \caption{NEL task: WER and frame-F1 scores on dev set}
    \label{fig:nel_corr_dev}
\end{figure}

Figure~\ref{fig:nel_corr_dev} shows the correlation between WER and frame-F1 on dev set. It follows a similar trend to test set (see Figure~\ref{fig:nel_corr}).

% \clearpage

%common appendix
\newpage
\section{Experiment detail}
Table~\ref{tab:nemo_modellist} shows NeMo model name list used in the experiment. Table~\ref{tab:models_size} shows the number of parameters for model used in the experiment.
%Please add the following packages if necessary:
%\usepackage{booktabs, multirow} % for borders and merged ranges
%\usepackage{soul}% for underlines
%\usepackage[table]{xcolor} % for cell colors
%\usepackage{changepage,threeparttable} % for wide tables
%If the table is too wide, replace \begin{table}[!htp]...\end{table} with
%\begin{adjustwidth}{-2.5 cm}{-2.5 cm}\centering\begin{threeparttable}[!htb]...\end{threeparttable}\end{adjustwidth}
\begin{table}[hp!]\centering
\caption{NeMo model list used in the experiment}\label{tab:nemo_modellist}
\resizebox{10cm}{!}{%
\begin{tabular}{lrrrrr}\toprule
NeMo model &DAC &QA &SUMM &NEL \\\midrule
QuartzNet15x5Base-En &o &o &o &o \\
stt\_en\_citrinet\_1024 &o &o &o &o \\
stt\_en\_citrinet\_1024\_gamma\_0\_25 &o &o &o &o \\
stt\_en\_citrinet\_256 &o &o &o &o \\
stt\_en\_citrinet\_256\_gamma\_0\_25 &o &o &o &o \\
stt\_en\_citrinet\_512 &o &o &o &o \\
stt\_en\_citrinet\_512\_gamma\_0\_25 &o &o &o &o \\
stt\_en\_conformer\_ctc\_large &o &o &o &o \\
stt\_en\_conformer\_ctc\_large\_ls &o &o &o &o \\
stt\_en\_conformer\_ctc\_medium &o &o &o &o \\
stt\_en\_conformer\_ctc\_medium\_ls &o &o &o &o \\
stt\_en\_conformer\_ctc\_small &o &o &o &o \\
stt\_en\_conformer\_ctc\_small\_ls &o &o &o &o \\
stt\_en\_conformer\_ctc\_xlarge &o &o &o &o \\
stt\_en\_conformer\_transducer\_large &o &o &o &o \\
stt\_en\_conformer\_transducer\_large\_ls &o &o &o &o \\
stt\_en\_conformer\_transducer\_medium &o &o &o &o \\
stt\_en\_conformer\_transducer\_small &o &o &o &o \\
stt\_en\_conformer\_transducer\_xlarge &o &o &o &o \\
stt\_en\_conformer\_transducer\_xxlarge &o &o &o &o \\
stt\_en\_contextnet\_1024 &o &o &o &o \\
stt\_en\_contextnet\_1024\_mls &o &o &o &o \\
stt\_en\_contextnet\_256 &o &o &o &o \\
stt\_en\_contextnet\_256\_mls &o &o &o &o \\
stt\_en\_contextnet\_512 &o &o &o &o \\
stt\_en\_contextnet\_512\_mls &o &o &o &o \\
stt\_en\_jasper10x5dr &o &o &o &o \\
stt\_en\_quartznet15x5 &o &o &o &o \\
stt\_en\_squeezeformer\_ctc\_large\_ls &o &o &o &o \\
stt\_en\_squeezeformer\_ctc\_medium\_large\_ls &o &o &o &o \\
stt\_en\_squeezeformer\_ctc\_medium\_ls &o &o &o &o \\
stt\_en\_squeezeformer\_ctc\_small\_ls &o &o &o &o \\
stt\_en\_squeezeformer\_ctc\_small\_medium\_ls &o &o &o &o \\
stt\_en\_squeezeformer\_ctc\_xsmall\_ls &o &o &o &o \\
\bottomrule
\end{tabular}
}
\end{table}

\begin{table}[hp!]\centering
\caption{Model parameter size used in experiment. We use \textit{base} sized model when there are multiple variants of the pre-trained model except off-the-shelf ASR model}\label{tab:models_size}
\begin{tabular}{llr}\toprule
Type &model name & parameter size \\\midrule
\multirow{3}{*}{Speech model} &wav2vec2 &95M \\
& DUAL (k-means model and Longformer part) & 149M \\
&TEDLIUM3-Conformer & 48.8M\\
&Hubert-ASR (Conformer part excluding Hubert)& 49.1M\\ 
&W2V2-ASR (Conformer part excluding wav2vec2)& 49.1M\\\midrule
Text model &DeBERTa &139M \\ \midrule
\multirow{35}{*}{off-the-shelf ASR model} &Whisper-en &71M \\
&QuartzNet15x5Base-En &18M \\
&stt\_en\_citrinet\_1024 &143M \\
&stt\_en\_citrinet\_1024\_gamma\_0\_25 &141M \\
&stt\_en\_citrinet\_256 &10M \\
&stt\_en\_citrinet\_256\_gamma\_0\_25 &9M \\
&stt\_en\_citrinet\_512 &36M \\
&stt\_en\_citrinet\_512\_gamma\_0\_25 &36M \\
&stt\_en\_conformer\_ctc\_large &121M \\
&stt\_en\_conformer\_ctc\_large\_ls &121M \\
&stt\_en\_conformer\_ctc\_medium &30M \\
&stt\_en\_conformer\_ctc\_medium\_ls &30M \\
&stt\_en\_conformer\_ctc\_small &13M \\
&stt\_en\_conformer\_ctc\_small\_ls &12M \\
&stt\_en\_conformer\_ctc\_xlarge &635M \\
&stt\_en\_conformer\_transducer\_large &120M \\
&stt\_en\_conformer\_transducer\_large\_ls &120M \\
&stt\_en\_conformer\_transducer\_medium &32M \\
&stt\_en\_conformer\_transducer\_small &14M \\
&stt\_en\_conformer\_transducer\_xlarge &644M \\
&stt\_en\_conformer\_transducer\_xxlarge &998M \\
&stt\_en\_contextnet\_1024 &144M \\
&stt\_en\_contextnet\_1024\_mls &144M \\
&stt\_en\_contextnet\_256 &14M \\
&stt\_en\_contextnet\_256\_mls &14M \\
&stt\_en\_contextnet\_512 &40M \\
&stt\_en\_contextnet\_512\_mls &40M \\
&stt\_en\_jasper10x5dr &332M \\
&stt\_en\_quartznet15x5 &18M \\
&stt\_en\_squeezeformer\_ctc\_large\_ls &236M \\
&stt\_en\_squeezeformer\_ctc\_medium\_large\_ls &125M \\
&stt\_en\_squeezeformer\_ctc\_medium\_ls &77M \\
&stt\_en\_squeezeformer\_ctc\_small\_ls &18M \\
&stt\_en\_squeezeformer\_ctc\_small\_medium\_ls &28M \\
&stt\_en\_squeezeformer\_ctc\_xsmall\_ls &9M \\
\bottomrule
\end{tabular}
\end{table}

\end{document}